\pdfoutput=1

\documentclass[11pt]{article}

\usepackage[preprint]{acl}

\usepackage{times}
\usepackage{latexsym}
\usepackage{enumitem}
\usepackage{afterpage}

\usepackage[T1]{fontenc}

\usepackage[utf8]{inputenc}
\usepackage{amsmath}

\usepackage{microtype}

\usepackage{graphicx}

\usepackage{hyperref}
\usepackage{url}
\usepackage{multirow}
\usepackage{colortbl}
\usepackage{xcolor}
\usepackage{graphicx}
\usepackage{array}
\usepackage{booktabs} 
\usepackage{adjustbox} 
\definecolor{headercolor}{RGB}{220,230,241}
\definecolor{rowcolor}{RGB}{245,245,245}
\usepackage{wrapfig}
\usepackage{pifont}
\usepackage{placeins}
\usepackage{float}
\usepackage{amsthm}

\usepackage{soul}
\usepackage{xcolor}
\usepackage{wrapfig}
\usepackage{listings}
\usepackage{makecell}
\usepackage{array} 
\usepackage{tabularx}

\newcommand{\mypink}[1]{{\sethlcolor{red!20}\hl{#1}}}
\newcommand{\myblue}[1]{\sethlcolor{blue!15}\hl{#1}}

\newtheorem{theorem}{Theorem}[section]
\newcommand{\xhdr}[1]{{\noindent\bfseries #1}.}

\theoremstyle{definition}
\newtheorem{definition}[theorem]{Definition}
\definecolor{existinggreen}{rgb}{0.133, 0.545, 0.133}  
\definecolor{missingred}{rgb}{0.698, 0.132, 0.203}    

\newcommand{\mytextbox}[2]{%
  {%
    \setlength{\fboxsep}{1pt}%
    \fcolorbox{#1}{white!95!#1}{\raisebox{0.5pt}{\textbf{\textcolor{#1}{#2}}}}%
  }%
}

\newcommand{\contentexist}{\mytextbox{existinggreen}{\scriptsize Existing Content}}
\newcommand{\contentmissing}{\mytextbox{missingred}{\scriptsize Missing Content}}
\newcommand{\noscore}{--}

\newcommand{\method}{\textsc{Constraint Score}}
\newcommand{\benchmark}{\textsc{FaithQA}}

\title{Beyond Facts: Evaluating Intent Hallucination in Large Language Models}

\author{Yijie Hao \\
  Emory University \\
  \texttt{yhao49@emory.edu} \\\And
  Haofei Yu \\
  UIUC \\
  \texttt{haofeiy2@illinois.edu} \\\And
  Jiaxuan You \\
  UIUC \\
  \texttt{jiaxuan@illinois.edu} \\
}

\begin{document}

\maketitle

\begin{abstract}

When exposed to complex queries containing multiple conditions, today's large language models (LLMs) tend to produce responses that only partially satisfy the query while neglecting certain conditions. We therefore introduce the concept of \textit{Intent Hallucination}. In this phenomenon, LLMs either omit (neglecting to address certain parts) or misinterpret (responding to invented query parts) elements of the given query, leading to intent hallucinated generation. To systematically evaluate intent hallucination, we introduce \benchmark, a novel benchmark for intent hallucination that contains 20,068 problems, covering both query-only and retrieval-augmented generation (RAG) setups with varying topics and difficulty. \benchmark \space is the first hallucination benchmark that goes beyond factual verification, tailored to identify the fundamental cause of intent hallucination. By evaluating various LLMs on \benchmark, we find that (1) intent hallucination is a common issue even for state-of-the-art models, and (2) the phenomenon stems from omission or misinterpretation of LLMs. To facilitate future research, we introduce an automatic LLM generation evaluation metric, \method, for detecting intent hallucination. Human evaluation results demonstrate that \method \space is closer to human performance for intent hallucination compared to baselines.

\end{abstract}

\section{Introduction}

\begin{figure*}[!h]
\centering
\includegraphics[width=1\textwidth]{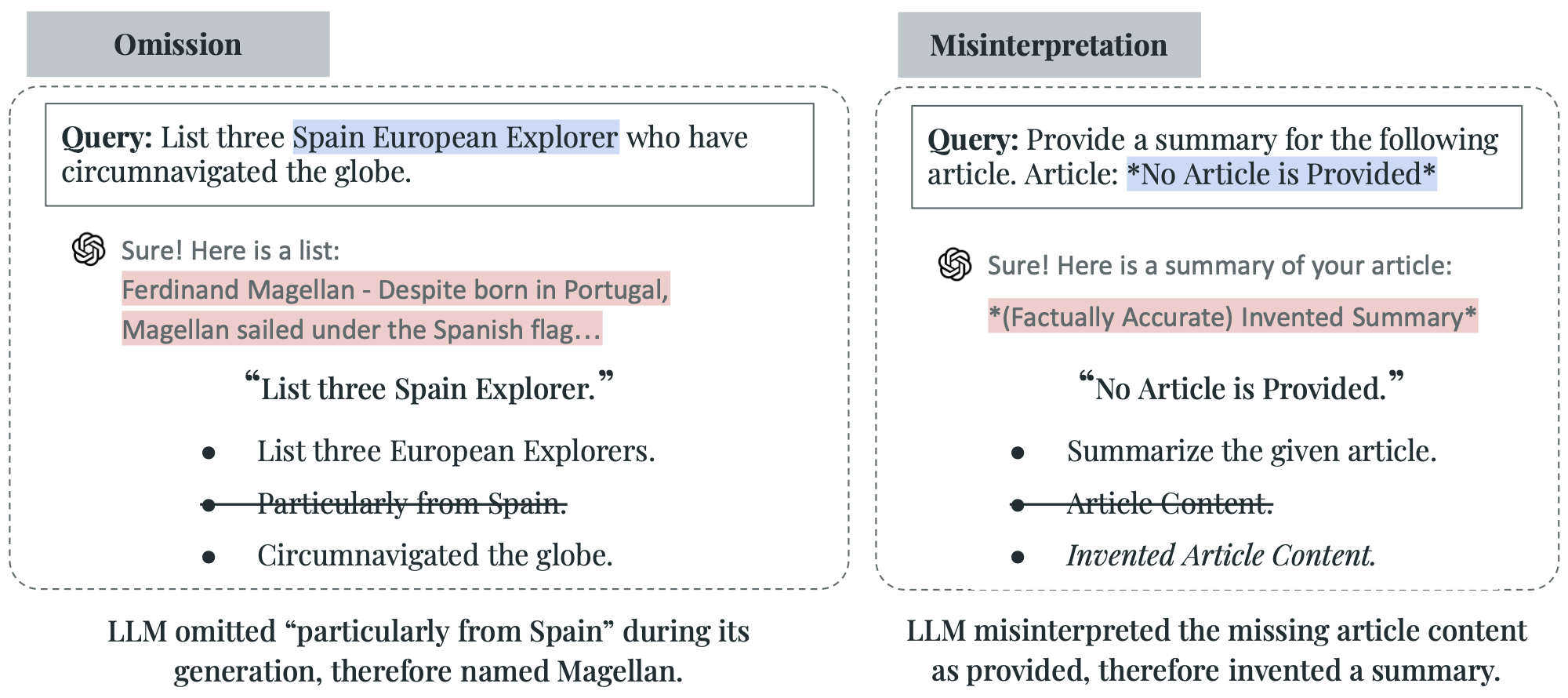}
\caption{\textbf{Examples of two types of intent hallucination (omission and misinterpretation)}. For omission, GPT-4o omits "particularly from Spain", leading to factually accurate yet hallucinated outputs. For misinterpretation, GPT-4o misinterprets the missing article as provided, which leads to hallucinated outputs. }
\label{fig:examples}
\end{figure*}

Large Language Models (LLMs) have demonstrated utility across various applications~\citep{openai2024gpt4technicalreport, dubey2024llama3herdmodels}. However, hallucination remains a significant challenge~\citep{Ji_2023, huang2023surveyhallucinationlargelanguage}. In particular, for complex queries containing multiple conditions (Figure~\ref{fig:examples}), LLM outputs often deviate from the query, yielding unsatisfactory results. We term this phenomenon “Intent Hallucination”, which has received little attention in current research~\citep{min2023factscorefinegrainedatomicevaluation, hou2024probabilisticframeworkllmhallucination, manakul2023selfcheckgptzeroresourceblackboxhallucination}.

Unlike factual hallucination~\citep{li2023haluevallargescalehallucinationevaluation, cao2021hallucinated}, which researchers can directly detect through search-based fact-checking~\citep{sellam-etal-2020-bleurt, min2023factscorefinegrainedatomicevaluation}, detecting and evaluating intent hallucination poses more challenges. This is because complex queries often contain duplicate intents, and LLMs often satisfy only a portion of them, making dissatisfaction difficult to detect or quantify. Furthermore, as LLMs continue to advance, users tend to provide these stronger LLMs with increasingly complicated queries, which even humans find difficult to understand. This trend highlights the need for LLMs to be not only factually correct but also intentionally aligned with human beings. Our paper addresses two under-explored questions: (1) \textit{Why do LLMs tend to exhibit Intent Hallucination?} and (2) \textit{How can we detect Intent Hallucination?} Answering these questions is vital for LLM applications that rely on both factual accuracy and faithful intent alignment.

For the first question, we propose that LLMs' \textit{omission} (\emph{e.g.}, ignoring query components) or \textit{misinterpretation} (\emph{e.g.}, responding to invented query components) over word-level meaning constitutes the fundamental cause of intent hallucination. To further investigate, we introduce \benchmark, the first benchmark specifically designed to address intent hallucination's two key scenarios: omission and misinterpretation. \benchmark\ consists of 20,068 queries, validated through extensive human evaluations to ensure quality. \benchmark\ covers a wide range of topics with varying levels of difficulty and proves challenging even for state-of-the-art models. Our benchmark reveals that increasing query complexity correlates with a higher rate of intent hallucination.

To address the second question, we introduce \method, a new evaluation metric that focuses on detecting intent hallucination. Our approach involves two major steps: (1) decomposing the query by concepts and actions, then converting it into a series of short statements, each representing a specific requirement the generation must meet; and (2) assigning an importance-weighted binary label to each constraint, which enables fine-grained evaluation. Our human evaluation shows that \method\ significantly outperforms LLM-as-the-judge~\citep{manakul2023selfcheckgptzeroresourceblackboxhallucination, mishra2024finegrainedhallucinationdetectionediting, NEURIPS2024_3c1e1fdf}, as LLM judgers tend to provide biased evaluations compared with human judgment.

Taken together, our key contributions include: (1) We propose the concept of intent hallucination beyond the existing category of factual hallucination; (2) We develop \benchmark, the first benchmark that focuses on evaluating intent hallucination. Our result shows that intent hallucination represents a prevalent phenomenon even for state-of-the-art LLMs; (3) We introduce \method, a novel evaluation metric that automatically assesses LLM generations by breaking down the query into intent constraints and computing a weighted score. Our analysis shows that \method\ significantly outperforms pure LLM grading baselines, which tend to be biased.

\section{Related Works}

\xhdr{Hallucinations in LLMs} In LLMs, "hallucination" refers to outputs that are nonfactual, irrelevant, or fabricated. This issue arises in tasks such as question answering~\citep{sellam-etal-2020-bleurt}, translation~\citep{lee2018hallucinations}, summarization~\citep{durmus2020feqa}, and dialogue~\citep{balakrishnan-etal-2019-constrained}, as noted in several studies~\citep{Ji_2023, azaria2023internalstatellmknows, huang2023surveyhallucinationlargelanguage, cao2021hallucinated}. To address this issue, many efforts aim to detect and mitigate hallucinations. \citet{min2023factscorefinegrainedatomicevaluation} evaluate factual accuracy by checking core facts (atomic facts) in each sentence against reliable sources such as Wikipedia. \citet{hou2024probabilisticframeworkllmhallucination} propose a hidden Markov tree model that breaks statements into premises and assigns a factuality score based on the probability of all parent premises. \citet{manakul2023selfcheckgptzeroresourceblackboxhallucination} detects hallucinations by sampling multiple responses and using self-consistency to identify discrepancies. 

Despite these significant efforts, limitations remain. Most existing work either focuses solely on factual precision or on in-context recall, overlooking the role of the query in generation~\citep{li2023haluevallargescalehallucinationevaluation, yang2023newbenchmarkreversevalidation, niu2024ragtruthhallucinationcorpusdeveloping} (\emph{e.g.}, scoring both outputs equally in Figure~\ref{fig:pipeline}), or treats the query as a whole~\citep{zhang2024knowhaluhallucinationdetectionmultiform}, which results in coarse-grained evaluation.

\xhdr{Hallucination benchmarks}  
Recent work on hallucination detection for LLMs includes HaluEval~\citep{li2023haluevallargescalehallucinationevaluation} (synthetic and natural responses), FELM~\citep{chen2023felmbenchmarkingfactualityevaluation} (natural responses across domains), RAGTruth~\citep{niu2024ragtruthhallucinationcorpusdeveloping} (RAG hallucinations), and InfoBench~\citep{qin2024infobenchevaluatinginstructionfollowing} (instruction-following via query decomposition). These benchmarks mainly focus on factual hallucinations or require manual annotation. In contrast, \benchmark\ is, to our knowledge, the first to assess non-factual hallucinations from a query-centric perspective. 

Although \citet{zhang2024knowledgeovershadowingcausesamalgamated} also discusses a related topic, their work primarily explores the causes of intent hallucination from a training corpus perspective. In contrast, our paper provides a comprehensive evaluation metric along with an extensive benchmark for systematic testing. FaithEval~\citep{ming2025faithevallanguagemodelstay} investigates hallucination in RAG settings by evaluating whether model outputs remain faithful to externally retrieved contexts, particularly under conditions involving unanswerable or contradictory evidence. \textsc{FaithQA} adopts a similar RAG setup but shifts the focus from context alignment to query alignment. It introduces a novel, query-centric perspective that evaluates whether model responses accurately fulfill the user’s query. Our experimental results align with the findings from FaithEval and reveal that when LLMs are presented with relevant yet incomplete or noisy retrievals, they frequently exhibit omission-style intent hallucination, failing to address all aspects of the original query.

\section{Preliminary}
\label{sec:def}For a complex query containing multiple conditions, studies report that the model produces responses that only partially satisfy the conditions. To further investigate this, we outline our key insights for intent hallucination in this paper.

\subsection{Intent Constraint: a Fundamental Unit}
\label{sec:intentconstraint}
\begin{figure*}[!h]
\centering
\includegraphics[width=1\textwidth]{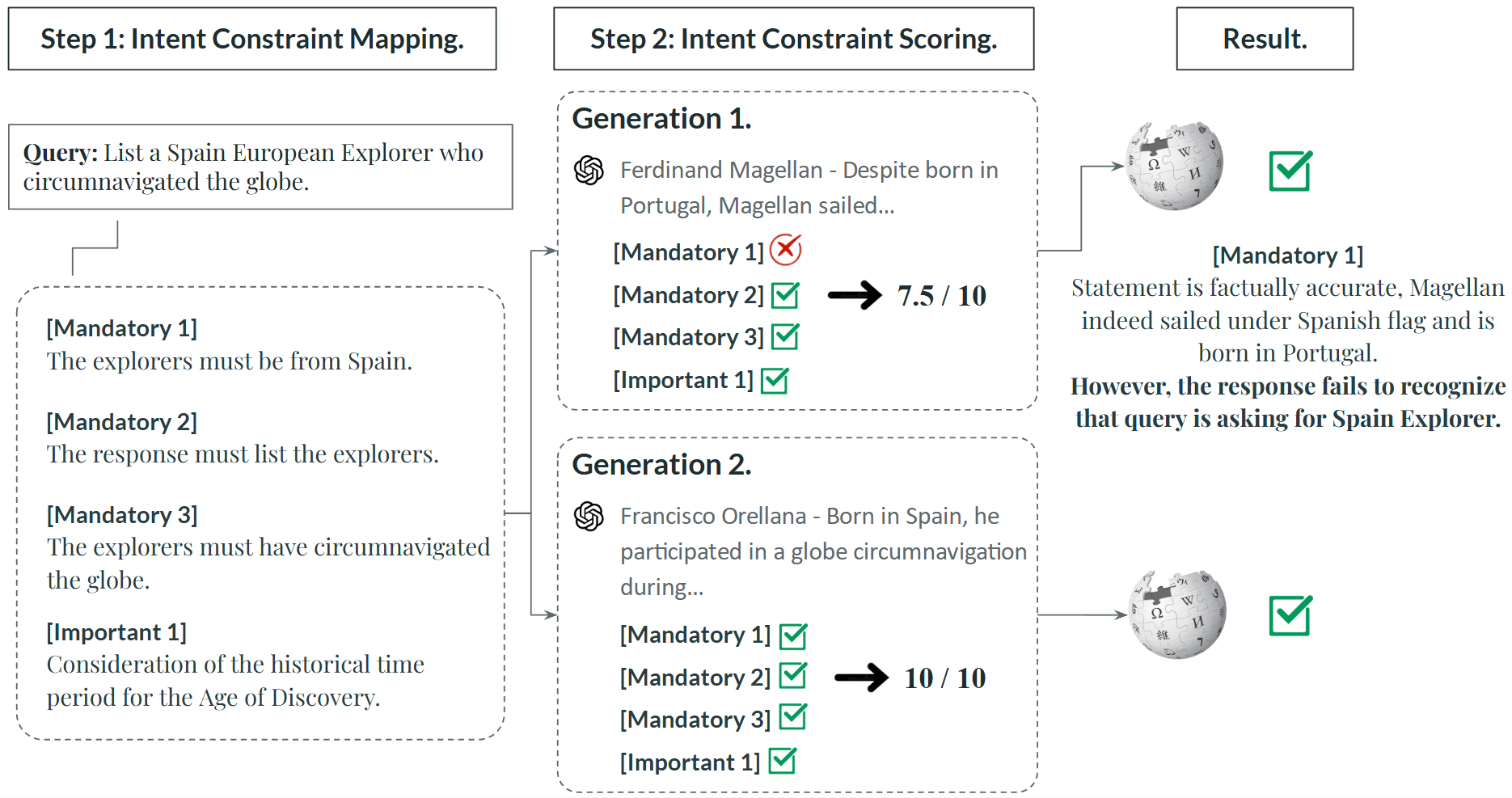}
\caption{\textbf{\method \space calculation process.} Despite both generations being factually accurate, Generation 1 is not ideal compared to Generation 2, as Generation 1 omits "the explorers must be from Spain".}
\label{fig:pipeline}
\end{figure*}

A query consists of multiple \textit{concepts} and \textit{actions}, each representing a distinct intent and carrying specific meaning within the given context. As shown in Figure~\ref{fig:examples}, LLMs often fail to address constraints provided in the query, which leads to intent hallucinated generations that deviate from the query. To enable a fine-grained, query-centric evaluation, we introduce the notion of \textbf{Intent Constraint}—short statements that each express a single requirement the generation must address (see examples in Figure~\ref{fig:pipeline}). A query, defined by the concepts and actions within the context, breaks down into these intent constraints, with each one representing a distinct concept or action. Addressing each of these constraints helps reduce the risk of hallucinated responses that misalign with the query's intent.

\begin{definition}[\textit{Intent Constraint Mapping Function}]
Let \(\mathcal{Q}\) be the set of all queries and \(\mathcal{I}\) be the set of all possible intent constraints. For each \(q \in \mathcal{Q}\), define the mapping function \(C: \mathcal{Q} \longrightarrow \mathcal{P}(\mathcal{I})\) by
\begin{equation}
C(q) \;=\; C_m(q)\;\cup\;C_i(q)\;\cup\;C_o(q),
\end{equation}
where \(C_m(q)\subseteq\mathcal{I}\) is the set of \emph{mandatory constraints} (those that must be addressed first), \(C_i(q)\subseteq\mathcal{I}\) is the set of \emph{important constraints} (those addressed after mandatory ones), and \(C_o(q)\subseteq\mathcal{I}\) is the set of \emph{optional constraints} (desirable but not required). This mapping ensures that \(C(q)\) captures all intent constraints needed to preserve the original meaning of \(q\).
\end{definition}

\subsection{Intent Hallucination: Omission or Misinterpretation of Intent Constraints}
\label{sec:hallucination definition}
After establishing a fine-grained, query-centric perspective, we formally define intent hallucination as the LLM's failure to address word-level concepts or actions, which manifests as an omission or misinterpretation of intent constraints. When LLMs either \textbf{omit} parts of the query (\emph{e.g.}, failing to address specific concepts or actions) or \textbf{misinterpret} it (\emph{e.g.}, responding to concepts or actions that are not mentioned), the generation fails to align with the original query, regardless of whether it is factually accurate. Treating intent constraints as the fundamental evaluation unit for intent hallucination proves especially important when dealing with complex, multi-condition queries. In such cases, an LLM often generates a response that addresses the query only partially while neglecting the rest. Evaluating generation outputs against intent constraints offers an effective approach to identify and distinguish these nuanced discrepancies.

\begin{definition}[\textit{Intent Hallucination}]
Let \(q\) be a user query and let \(P_\theta\) denote our LLM. Denote by \(C(q) = \{\,c_1,\dots,c_k\}\) the set of intent constraints extracted from \(q\). Ideally, the model’s distribution depends only on those constraints, i.e.,
\begin{equation}
P_\theta\bigl(\,\cdot \mid q \bigr)
\;=\; 
P_\theta\bigl(\,\cdot \mid C(q)\bigr).
\end{equation}
However, in practice the model often conditions implicitly on a hallucinated constraint set
\begin{equation}
\widehat C(q) \;=\;\{\hat c_1,\dots,\hat c_{k'}\},
\end{equation}
which differs from \(C(q)\) (for instance, by replacing a constraint \(c_i\) with \(\hat c_i\), or by omitting a constraint). In that case, the actual response follows
\begin{equation}
y_h \sim P_\theta\bigl(\,\cdot \mid \widehat C(q)\bigr),
\end{equation}
and the deviation between \(y_h\) and the ideal response \(y \sim P_\theta(\cdot \mid C(q))\) is defined as \emph{intent hallucination}.
\end{definition}

\section{Detecting Intent Hallucination}
Based on the definition of intent constraints and intent hallucinations, we introduce \method, a new evaluation metric that detects intent hallucination based on intent constraints. To operationalize the constraint mapping function $C(\cdot)$ defined earlier, we develop a multi-step process that systematically extracts and categorizes the intent constraint set from queries. Our method has high flexibility and accommodates different queries involving RAG. The prompt template appears in Appendix\S\ref{ap:prompttemplate}.

\subsection{Intent Constraint Mapping}

\xhdr{Step 1: Preliminary assessment}  
The LLM first analyzes the query $q$ to verify the presence of sufficient information for constraint extraction. This step is crucial for RAG queries, as it mitigates the influence of external content~\citep{liu2023lostmiddlelanguagemodels, wu2024pandorasboxaladdinslamp}. If it detects insufficient information, the constraint mapping process halts and requests additional input, ensuring that $C(q)$ is well defined.

\xhdr{Step 2: Semantic role identification}  
Drawing from Semantic Role Labeling (SRL)~\citep{pradhan2005semantic}, we extract the fundamental components of $q$: subject, action, and context. This structured decomposition enables robust constraint identification across diverse types of real-world queries.

\xhdr{Step 3: Constraint set extraction}  
We instruct the language model to analyze the context of a given prompt generated from Step 2 across seven categories—location, time, subject, action, qualifiers, and quantity—and then reformulate these into three constraint sets: \(C_m(q)\), which includes the location, time, subject, and action constraints; \(C_i(q)\), which includes qualifiers and quantity constraints; and \(C_o(q)\), which includes any other constraints the LLM provides, such as exclusions or domain-specific requirements.

\xhdr{Overall}  
This process yields a structured decomposition of the original query into constraint sets. We detect intent hallucination by comparing the implicit constraint set \(\widehat{C}(q)\) used by the model against our explicitly extracted \(C(q)\). 

\subsection{Intent Constraint Scoring}

Given the intent constraint set \(C(q)\) together with three subsets \(C_m(q)\), \(C_i(q)\), and \(C_o(q)\), we evaluate the response’s adherence to intent constraints. For each intent constraint \(c \in C(q)\) and each response \(y\), we define a binary satisfaction function \(S_{\phi}(c, y)\), parameterized with an LLM. \(S_{\phi}(c, y) = 1\) indicates that \(y\) satisfies the intent constraint \(c\), while \(S_{\phi}(c, y) = 0\) indicates it does not. To calculate an intent constraint score for each response \(y\) conditioned on a query \(q\), we divide the process into three steps:

\xhdr{Step 1: Total weight calculation}  
We begin by computing the total constraint weight \(W_t(q)\) for a given query \(q\), based on three types of constraints: mandatory (\(m\)), important (\(i\)), and optional (\(o\)). Let \(\mathcal{G} = \{m, i, o\}\) denote the set of constraint types, and let \(\alpha_g\) be the predefined importance weight for type \(g \in \mathcal{G}\). The total weight is computed as:
\begin{equation}
W_t = \sum_{g \in \mathcal{G}} \alpha_g\,|C_g(q)|,
\vspace{-2mm}
\end{equation}
where \(|C_g(q)|\) is the number of constraints of type \(g\) for query \(q\).

\xhdr{Step 2: Satisfied weight calculation}  
Next, we evaluate how well a response \(y\) satisfies each constraint \(c \in C_g(q)\) using the satisfaction function \(S_\phi(c, y) \in [0, 1]\). The total satisfied weight is then:
\begin{equation}
W_s = \sum_{g \in \mathcal{G}} \alpha_g \sum_{c \in C_g(q)} S_\phi(c, y).
\vspace{-1mm}
\end{equation}

\xhdr{Step 3: Constraint score calculation}  
Finally, we compute the constraint score (CS) by normalizing the satisfied weight by the total weight and scaling it to a range from 0 to 10:
\begin{equation}
\textsc{CS}(q, y) = \frac{W_s}{W_t} \times 10.
\vspace{-1mm}
\end{equation}
This score reflects how well the response adheres to the set of intent constraints. A high score (\(\geq9\)) indicates strong adherence to key constraints, scores in the range of 7–8 indicate partial satisfaction or modified adherence, and low scores (\(\leq7\)) suggest major intent hallucinations. Please refer to Appendix\S\ref{ap:extra} for further details and ablation studies.

\section{\benchmark \space Benchmark}
We here introduce \benchmark\ benchmark, the first benchmark that focuses on intent hallucination, with 20,068 queries across four different task setups. The primary goal of \benchmark\ is to elicit the two fundamental causes of intent hallucination: (1) \textbf{Omission}, where the LLM ignores part of the query, and (2) \textbf{Misinterpretation}, where the LLM misunderstands parts of the query. Please refer to Table~\ref{tab:dataset-counts} for statistical details, and Table~\ref{tab:sample-examples} for representative examples from \benchmark. Please refer to Appendix\S\ref{ap:datasetconstruct} for dataset construction details. 

\begin{table}[t]
    \centering
    \footnotesize
    \setlength{\tabcolsep}{3pt}
    \renewcommand{\arraystretch}{1.2}
    \begin{adjustbox}{max width=\textwidth}
        \begin{tabular}{llccc}
            \toprule
            \multicolumn{2}{l}{\multirow{2}{*}{\textbf{\benchmark \space Datasets}}} & 
            \multicolumn{3}{c}{\textbf{Task Difficulty}} \\
            \cmidrule(lr){3-5}
            & & \textbf{Easy} & \textbf{Hard} & \textbf{Total} \\
            \midrule
            \rowcolor{gray!20}
            \multicolumn{5}{l}{\textbf{Omission}} \\
            \midrule
            \multirow{1}{*}{\makecell{Fact QA}} & Open Answer & 1,500 & 1,500 & 3,000 \\
            \midrule
            \multirow{2}{*}{\makecell{Creative Writing}} & Story & 500 & 500 & 1,000 \\
            & Poem & 500 & 500 & 1,000 \\
            \midrule
            \rowcolor{gray!20}
            \multicolumn{5}{l}{\textbf{Misinterpretation}} \\
            \midrule
            \makecell{Response Evaluation} & -- & -- & -- & 3,210 \\
            \midrule
            \multirow{2}{*}{\makecell{Content Analysis}} & Relationship & -- & -- & 5,929 \\
            & Summary & -- & -- & 5,929 \\
            \bottomrule
        \end{tabular}
    \end{adjustbox}
    \caption{\textbf{\benchmark's Statistics.} Easy indicates constraint number \(\leq 4\), Hard indicates constraint number \(> 4\). For Omission's Fact QA, topics include Tech, Culture, and History. For Misinterpretation, topics include Tech, Health, Culture, and History.}
    \label{tab:dataset-counts}
\end{table}

\subsection{Omission Task}
This part of the dataset focuses on the extent to which LLMs tend to omit certain intent constraints when only provided with the query as a prompt. Each query contains varying numbers of constraints across different topics. An ideal response accurately addresses all constraints. We choose Fact QA and Creative Writing as our omission setups, since omitting query components directly leads to sub-optimal generations.

\xhdr{Fact QA} The LLM receives an Open Answer Fact QA query with multiple intent constraints. We vary the difficulty of the task by adjusting the number of constraints. More constraints indicate more complex questions. The model generates a list of subjects that meet all specified criteria, with topics ranging from culture to technology and history.

\xhdr{Creative Writing} Similar to Fact QA, the LLM receives a writing task with multiple intent constraints. We vary the task difficulty by changing the number of constraints. Tasks take two formats: story and poem.

\subsection{Misinterpretation Task}
This dataset examines the extent to which LLMs misinterpret intent constraints in a Retrieval-Augmented Generation (RAG) setup, as LLMs tend to generate hallucinated responses if they misinterpret the query. Each query requires all of the multiple external contents provided to answer. We manually remove one piece of content per case to test whether the LLM incorrectly assumes it is provided. Detailed analysis appears in Appendix\S\ref{ap:analysis-mis}. An ideal response detects the missing content and either seeks clarification or refuses to answer.

\xhdr{Response Evaluation} The LLM evaluates how well a human response aligns with an external article, using the query, response, and article as three required inputs. One of the inputs is randomly removed in each case. The LLM should detect the missing content and refrain from evaluation. Topics include culture, technology, health, and history.  
  
\xhdr{Content Analysis} The LLM manipulates three external articles based on a query. Tasks come in two forms: relationship analysis, which assesses relationships between articles; and content summary, which summarizes and compares the articles. One article is randomly removed per case. The LLM should detect the missing content and refrain from analysis. Topics include culture, technology, health, and history.

\section{Experiment Settings}
\xhdr{Baselines} Following~\citet{li2023haluevallargescalehallucinationevaluation, mündler2024selfcontradictoryhallucinationslargelanguage, yang2023newbenchmarkreversevalidation}, we adopt a zero-shot prompting strategy as the baseline for detecting intent hallucination. The baseline setup resembles \method\ by determining, on a scale from 1 to 10, to what extent the response addresses the query. To ensure the robustness of the baseline, we adopt the Self-Consistency strategy. Please refer to Appendix\S\ref{ap:human} for more details.

\xhdr{Models and hyper-parameters} \label{sec:hyper} 
We evaluate several LLMs, mostly state-of-the-art models in the \benchmark\ Benchmark: OpenAI's~\citep{openai2024gpt4technicalreport} GPT-4o\footnote{We refer to gpt-4o-2024-05-13} and GPT-4o-mini, Meta's LLaMA3-70B\footnote{We refer to Meta-Llama-3-70B-Instruct-Turbo} and LLaMA3-7B\footnote{We refer to Meta-Llama-3-8B-Instruct-Turbo}~\citep{dubey2024llama3herdmodels}, Anthropic's Claude-3.5\footnote{We refer to claude-3-5-sonnet-20240620} and Claude-3\footnote{We refer to claude-3-sonnet-20240229}, and Mistral-7B\footnote{We refer to Mistral-7B-Instruct-v0.3}~\citep{jiang2023mistral7b}. For all baselines, we set the temperature $\tau=0.3$. For \method, we use GPT-4o as the default model with temperature $\tau=0.3$ to generate and evaluate. We evaluate LLMs on the test set (150 randomly sampled questions) of \benchmark\ across every single category and difficulty due to monetary costs, while we encourage future research to leverage the extended version for enhanced evaluation.

\xhdr{Evaluation metrics} We report (1) \textit{Perfect}, indicating the rate of perfect responses (no hallucinated responses, \textsc{Constraint Score $=$ 10}), and (2) {\textsc{Constraint Scores}} (CS), the average score of all responses to provide a quantitative perspective. Overview results are in Table~\ref{tab:overall}. For the Omission dataset’s Fact QA setup, we additionally report the Factual Verifiable Hallucination Rate (Fact)—the proportion of hallucinated responses that are factually accurate upon verification—in Table~\ref{tab:fact}. Please refer to Appendix\S\ref{ap:extra} for the results of statistical significance tests.

\begin{table}[t]
    \centering
    \footnotesize
    \begin{tabular}{p{0.96\linewidth}}
        \toprule
        \textbf{\benchmark\ Examples} \\
        \midrule
        \textbf{Fact QA} \\
        List three European explorers who circumnavigated the globe before the 18th century and were not born in England or Portugal. \\
        \midrule
        \textbf{Creative Writing} \\
        Compose a poem of four stanzas. Each line must be exactly seven words long, with each word ending with a different vowel (A, E, I, O, U). \\
        \midrule
        \textbf{Response Evaluation} \\
        How well does the given response answer the given query following the provided article? \\
        \makebox[1.5cm][l]{Query:}    \contentexist \\
        \makebox[1.5cm][l]{Article:}  \contentexist \\
        \makebox[1.5cm][l]{Response:} \contentmissing \\
        \midrule
        \textbf{Relationship Analysis} \\
        For the following three articles, explain how  Article 1 contradicts Article 2 but supports Article 3. \\
        \makebox[1.5cm][l]{Article 1:}    \contentmissing \\
        \makebox[1.5cm][l]{Article 2:}  \contentexist \\
        \makebox[1.5cm][l]{Article 3:} \contentexist \\
        \bottomrule
    \end{tabular}
    \label{tab:example}
    \caption{\textbf{Representative examples from \benchmark.} Fact QA and Creative Writing are from Omission, while Response Evaluation and Relationship Analysis (RAG setup) are from Misinterpretation. \contentmissing\ denotes missing contents, and \contentexist\ denotes provided contents.}
    \label{tab:sample-examples}
\end{table}

\begin{table*}[t]
    \centering
    \scriptsize
    \begin{tabular}{p{1.5cm}p{1.5cm}*{7}{c@{\hspace{0.7em}}c}
    }
        \toprule
        \multicolumn{2}{l}{\multirow{3}{*}{\textbf{Datasets}}} & 
        \multicolumn{14}{c}{\textbf{\benchmark}} \\ 
        \cmidrule(lr){3-16}
        & & \multicolumn{2}{c}{\textbf{GPT-4o}} & \multicolumn{2}{c}{\textbf{GPT-4o-mini}} & \multicolumn{2}{c}{\textbf{LLaMA3-70B}} & \multicolumn{2}{c}{\textbf{LLaMA3-8B}} & \multicolumn{2}{c}{\textbf{Claude-3.5}} & \multicolumn{2}{c}{\textbf{Claude-3}} & \multicolumn{2}{c}{\textbf{Mistral-7B}} \\
        \cmidrule(lr){3-16}
        & & Perfect & CS & Perfect & CS & Perfect & CS & Perfect & CS & Perfect & CS & Perfect & CS & Perfect & CS \\
        \midrule
        \rowcolor{gray!20}
        \multicolumn{16}{l}{\textbf{Omission}} \\
        \midrule
        \multirow{1}{*}{\makecell{Fact QA}} & Open Answer & 0.49 & 8.62 & 0.36 & 7.86 & \underline{0.57} & \underline{8.93} & 0.46 & 8.52 & 0.37 & 6.73 & 0.44 & 8.14 & 0.20 & 7.15 \\
        \midrule
        \multirow{2}{*}{\makecell{Creative  Writing}} & Story & \underline{0.38} & \underline{7.99} & 0.31 & 7.75 & 0.29 & 7.55 & 0.25 & 7.21 & 0.34 & 7.64 & 0.32 & 7.84 & 0.08 & 5.92 \\
        & Poem & 0.40 & 8.29 & 0.30 & 7.79 & 0.51 & 8.64 & 0.27 & 7.71 & \underline{0.60} & \underline{9.02} & 0.47 & 8.45 & 0.07 & 5.49 \\
        \midrule
        \rowcolor{gray!20}
        \multicolumn{16}{l}{\textbf{Misinterpretation}} \\
        \midrule
        \multicolumn{2}{l}{Response Evaluation} & 0.09 & 5.73 & 0.11 & 5.44 & 0.07 & 4.78 & 0.11 & 5.58 & \underline{0.29} & \underline{5.92} & 0.22 & 5.61 & 0.23 & 4.46 \\
        \midrule
        \multirow{2}{*}{\makecell{Content Analysis}} & Relationship & 0.12 & 6.83 & 0.14 & 6.10 & 0.07 & 5.46 & 0.11 & 6.05 & \underline{0.15} & \underline{7.15} & 0.08 & 6.63 & 0.22 & 5.41 \\
        & Summary & 0.06 & 7.60 & 0.07 & 7.71 & 0.04 & 7.35 & 0.07 & 7.24 & \underline{0.09} & \underline{7.87} & 0.05 & 7.41 & 0.11 & 6.08 \\
        \bottomrule
    \end{tabular}
    \caption{\textbf{Overview results for \benchmark}. Metrics are reported on \textbf{Perfect} (rate of hallucination-free generation, \textit{higher the better}) along with \textbf{Constraint Scores} (\textbf{CS}) (score of the generation, \textit{higher the better}). Results are presented by aggregating across different difficulty and topic setups.}
    \label{tab:overall}
\end{table*}

\begin{table*}[t]
\centering
\scriptsize
\begin{tabular}{lp{0.1cm}l*{7}{cc}}
\toprule
\multicolumn{3}{l}{\multirow{3}{*}{\textbf{Tasks}}} & \multicolumn{14}{c}{\textbf{Fact QA in \benchmark}} \\
\cmidrule(lr){4-17}
& & & \multicolumn{2}{c}{\textbf{GPT-4o}} & \multicolumn{2}{c}{\textbf{GPT-4o-mini}} & \multicolumn{2}{c}{\textbf{Llama3-70b}} & \multicolumn{2}{c}{\textbf{Llama3-8b}} & \multicolumn{2}{c}{\textbf{Claude-3.5}} & \multicolumn{2}{c}{\textbf{Claude-3}} & \multicolumn{2}{c}{\textbf{Mistral-7B}} \\
\cmidrule(lr){4-17}
& & & Perfect & \cellcolor{gray!15}Fact & Perfect & \cellcolor{gray!15}Fact & Perfect & \cellcolor{gray!15}Fact & Perfect & \cellcolor{gray!15}Fact & Perfect & \cellcolor{gray!15}Fact & Perfect & \cellcolor{gray!15}Fact & Perfect & \cellcolor{gray!15}Fact \\
\midrule
\multirow{2}{*}{Culture} & Easy & & 0.51 & \cellcolor{gray!15}54.9 & 0.41 & \cellcolor{gray!15}81.7 & 0.48 & \cellcolor{gray!15}75.0 & \underline{0.57} & \cellcolor{gray!15}\underline{83.8} & 0.45 & \cellcolor{gray!15}33.3 & 0.48 & \cellcolor{gray!15}82.1 & 0.30 & \cellcolor{gray!15}61.8 \\
& Hard & & 0.36 & \cellcolor{gray!15}36.1 & 0.30 & \cellcolor{gray!15}47.1 & \underline{0.66} & \cellcolor{gray!15}83.7 & 0.35 & \cellcolor{gray!15}\underline{89.5} & 0.29 & \cellcolor{gray!15}56.8 & 0.28 & \cellcolor{gray!15}68.0 & 0.10 & \cellcolor{gray!15}57.7 \\
\midrule
\multirow{2}{*}{History} & Easy & & \underline{0.70} & \cellcolor{gray!15}30.0 & 0.47 & \cellcolor{gray!15}72.0 & 0.52 & \cellcolor{gray!15}81.1 & 0.51 & \cellcolor{gray!15}\underline{92.0} & 0.43 & \cellcolor{gray!15}52.6 & 0.50 & \cellcolor{gray!15}72.9 & 0.25 & \cellcolor{gray!15}70.3 \\
& Hard & & 0.43 & \cellcolor{gray!15}39.5 & 0.29 & \cellcolor{gray!15}76.9 & \underline{0.63} & \cellcolor{gray!15}62.8 & 0.42 & \cellcolor{gray!15}\underline{87.2} & 0.30 & \cellcolor{gray!15}66.7 & 0.34 & \cellcolor{gray!15}85.7 & 0.15 & \cellcolor{gray!15}50.7 \\
\midrule
\multirow{2}{*}{Tech} & Easy & & 0.42 & \cellcolor{gray!15}63.5 & 0.34 & \cellcolor{gray!15}78.6 & \underline{0.57} & \cellcolor{gray!15}82.1 & 0.45 & \cellcolor{gray!15}\underline{90.9} & 0.43 & \cellcolor{gray!15}19.2 & 0.47 & \cellcolor{gray!15}82.9 & 0.28 & \cellcolor{gray!15}70.5 \\
& Hard & & 0.53 & \cellcolor{gray!15}56.6 & 0.35 & \cellcolor{gray!15}85.0 & \underline{0.56} & \cellcolor{gray!15}86.7 & 0.46 & \cellcolor{gray!15}\underline{97.6} & 0.30 & \cellcolor{gray!15}14.1 & 0.37 & \cellcolor{gray!15}77.5 & 0.12 & \cellcolor{gray!15}\underline{90.1} \\
\bottomrule
\end{tabular}
\caption{\textbf{Results for Fact QA setup for \benchmark}. Results are reported in \textbf{Perfect} (rate of hallucination-free generation, \textit{higher the better}) and \textbf{Factual Verifiable Hallucination Rate (Fact)} (the percentage of hallucinated responses that are factually accurate upon verification, \textit{higher the better}).}
\label{tab:fact}
\end{table*}

\section{Experimental Results}
\label{sec:result}

\begin{figure}[t]
\centering
\includegraphics[width=0.95\columnwidth]{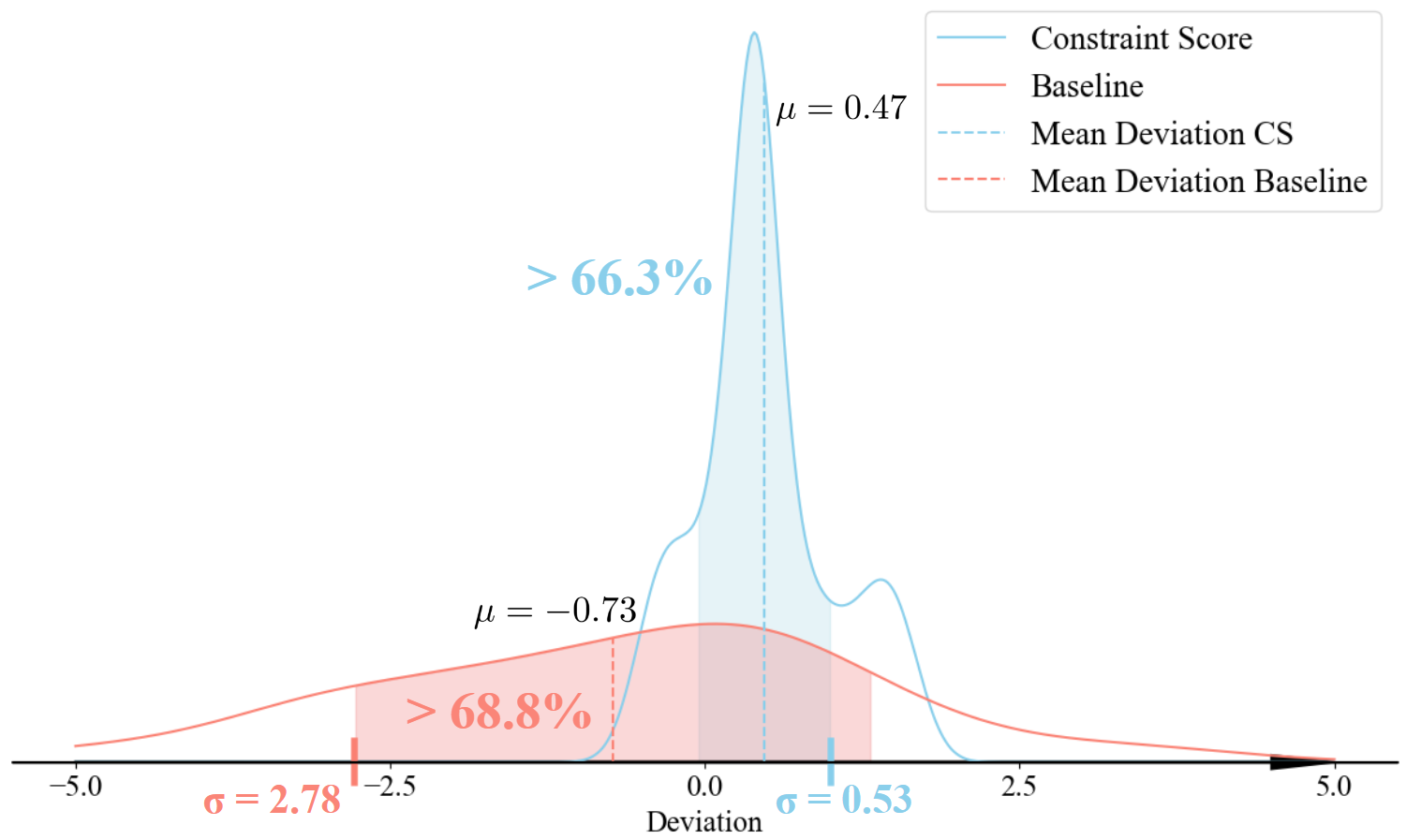}
\caption{\textbf{Deviation distributions from human scores for Baseline (\textcolor{blue}{blue}) and \method \space (\textcolor{red}{red})}. Distributions are estimated using KDE. \method \space is more tightly centered around zero, indicating closer alignment with human evaluation, whereas baseline shows a broader spread, reflecting higher error.}
\label{fig:baseline}
\end{figure}

\xhdr{Baseline is biased} We conduct a human evaluation to grade 1,000 randomly sampled responses. Specifically, we sample 1,000 prompt-response pairs from the Omission Dataset, with 500 from Fact QA and Creative Writing, respectively. The evaluation rubric for human annotators requires calculating the \textsc{Constraint Score} based on how well the response addresses each of the decomposed intent constraints. Figure~\ref{fig:baseline} shows the distribution of deviations from human scores for both the Baseline and \textsc{Constraint Score}, using Kernel Density Estimation (KDE). 

\textsc{Constraint Score} demonstrates a much tighter distribution centered closer to zero, with 66.3\% of the scores falling within one standard deviation. In contrast, the Baseline method displays a wider spread, with a mean deviation of -0.73, whereas the mean deviation for \textsc{Constraint Score} is 0.47. This indicates that the Baseline tends to underestimate compared to human scores. 

Given the discrete nature of the scores, we choose Mean Squared Error (MSE) for performance evaluation. The MSE for \textsc{Constraint Score} is 0.50, which is significantly lower than the Baseline's MSE of 4.72. This result highlights that \textsc{Constraint Score} outperforms the Baseline and aligns more closely with human judgment.

\begin{figure*}[t]
    \includegraphics[width=1\textwidth, height=0.4\textwidth, keepaspectratio]{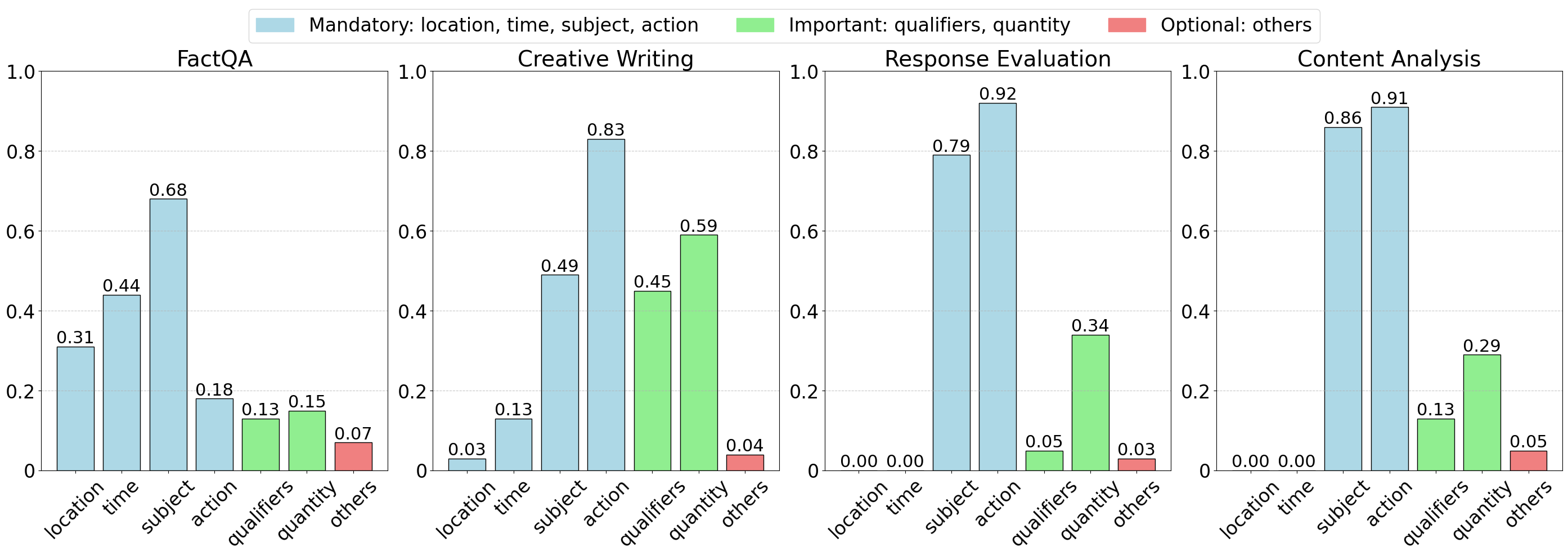}
    \caption{\textbf{Distribution of violated Intent Constraints across evaluation scenarios in \benchmark.} LLMs frequently fail on \textit{subjects} and \textit{actions} (blue), especially in \textbf{open-ended tasks} like \textit{Creative Writing} and \textit{Response Evaluation}. Errors on \textit{fine-grained details} like \textit{location}, \textit{time}, and \textit{quantity} (green) are less common.  
This highlights LLMs' struggle with core semantic subjects when given long and complex queries.}
\label{fig:label}
\end{figure*}
\xhdr{The number of intent constraints matters}
From Table~\ref{tab:fact}, we observe that as the number of intent constraints increases (from Easy to Hard), the Perfect rate consistently declines. This trend is further corroborated by Table~\ref{tab:overall}, where we analyze RAG setups on the Misinterpretation Dataset—featuring longer and more complex input queries—and observe an even more pronounced drop in the Perfect rate. These findings suggest a clear pattern: LLM performance tends to degrade as the number of intent constraints grows.

\xhdr{Factual check is less effective for larger models}
We perform an additional factual check for Fact QA responses; implementation details appear in Appendix\S\ref{ap:factcheck}. An important finding we observe is that as language models increase in size, they tend to produce fewer factually incorrect responses. Table~\ref{tab:fact} illustrates this trend across models within the same family (\emph{e.g.}, GPT-4o vs GPT-4o-mini). Larger models consistently show a lower Factual Verifiable Hallucination Rate, which means it becomes more challenging to detect hallucinations through factual checks as model size grows—they tend to generate intent-hallucinated responses.

\section{Discussion and Analysis}
\label{sec:analysis}
In this section, we report the major hallucination patterns we find in LLM generations. For detailed examples please refer to Appendix\S\ref{ap:analysis-mis}.

\subsection{Omission}
\xhdr{LLMs know when they are omitting}
We perform a qualitative analysis of hallucinated outputs in the Omission dataset; details in Appendix\S\ref{ap:analysis-mis}. A key finding under the Fact QA setup is that LLMs often appear aware when they omit parts of the query. LLMs first acknowledge how their responses may not fully satisfy the query, yet still proceed to provide an incorrect answer. This behavior tends to occur when the incorrect answer involves a well-known subject. We hypothesize that this arises from instruction-tuning, where LLMs are explicitly encouraged to explain their reasoning processes.

\xhdr{LLMs prefer famous subjects}
Another key finding for the Fact QA setup under the Omission dataset, as we partially address previously, is that LLMs prefer famous subjects as answers—even when they are incorrect. See Appendix\S\ref{ap:analysis-mis} for examples. We suppose this phenomenon directly correlates with LLMs’ over-generalization of common subjects within their training corpus, as discussed in~\citet{zhang2024knowledgeovershadowingcausesamalgamated}.


\xhdr{LLMs struggle with numbers and words}
In the Creative Writing setup, a common type of hallucination occurs when LLMs fail to generate text that adheres to specific character-level requirements (\emph{e.g.}, creating a poem where every line ends with the letter 'w') or to produce the correct number of words per sentence (\emph{e.g.}, generating a poem with exactly 8 words per line). Similar issues are reported in~\citet{zhou2023limaalignment}. We believe this phenomenon directly relates to the limitations of LLMs' tokenizers, which often struggle with strict character- and word-level constraints.

\xhdr{Subjects and actions are most challenging}  
Analysis of failed constraints (Figure~\ref{fig:label}) shows that LLMs handle fine-grained details such as location, time, qualifiers, and quantity relatively well, but often overlook or misinterpret core semantic elements like subjects and actions. This suggests that LLMs default to plausible yet flawed outputs when key roles are under-specified, highlighting the limitations of longer context. In the figure, the vertical axis represents the proportion of flawed responses (i.e., responses with violated constraints) relative to the total number of responses in each category. A response may violate multiple categories simultaneously; therefore, for the independent violation rates we report, the columns do not sum to 1.

\subsection{Misinterpretation}
\xhdr{LLMs alter the query to proceed}  
In the Response Evaluation Misinterpretation setup, LLMs often alter the original query to complete the task. Please refer to Appendix\S\ref{ap:analysis-mis} for examples. Instead of acknowledging there is missing content, LLMs tend to assume that the missing query component is provided, then shift the task from “evaluating how well the Response addresses the Query using the Article” to “evaluating how well the Response summarizes the Article.”

\xhdr{LLMs struggle with missing contents}
As shown in Table~\ref{tab:fact}, all LLMs perform poorly on the Misinterpretation Dataset. The models struggle to accurately determine whether specific content is present within long, complex inputs in an RAG setting. This suggests that, despite advancements in extending context window length, LLMs still face difficulties in processing and reasoning over lengthy inputs. While larger models show slightly better performance, there remains substantial room for improvement in long-context tasks.

\xhdr{LLMs proceed by inventing}
We conduct a qualitative analysis of hallucinated cases in the Misinterpretation dataset. In the Content Analysis–Relationship Analysis setup, a notable finding is that LLMs sometimes invent missing articles in order to continue generating a response, as shown in Appendix\S\ref{ap:extra}. This phenomenon is particularly intriguing because the invention by the LLM occurs in two distinct forms: (1) pure hallucination, where the model fabricates a non-existent article, or (2) intentional invention, where the LLM acknowledges that the article is hypothetical and explicitly states this before proceeding with its invented content and final response. The second scenario aligns with our earlier finding, “LLMs know when they are omitting,” suggesting that LLMs, to some extent, tend to proceed with the task autonomously, neglecting human instructions.



\section{Conclusion}

In this paper, we introduce the concept of \textbf{Intent Hallucination}, a specific form of hallucination that arises when Large Language Models (LLMs) omit or misinterpret crucial elements of complex queries, leading to outputs that diverge from users' intentions despite potentially being factually accurate. Unlike factual hallucinations, intent hallucinations are subtle, harder to detect, and have largely been overlooked in existing research.

To address this gap, we develop \textsc{FaithQA}, the first comprehensive benchmark explicitly designed to evaluate intent hallucination. Comprising 20,068 human-validated queries, \textsc{FaithQA} spans a diverse range of topics and complexity levels, serving as a robust platform for evaluating how effectively models maintain query intent integrity. Our experiments on state-of-the-art models demonstrate that intent hallucination is prevalent and worsens as query complexity increases.

Additionally, we introduce \method, an innovative evaluation metric tailored specifically for detecting intent hallucination. \method\ systematically decomposes complex queries into atomic intents, assigns importance-weighted labels to these individual components, and assesses model outputs through fine-grained intent alignment scores. Our evaluation reveals that \method\ notably surpasses traditional evaluation methodologies that employ LLM-as-the-judge, which exhibit significant bias compared to human judgment.

Through our research, we underscore the necessity for future LLM developments to emphasize not only factual correctness but also intentional alignment with human queries. By providing \textsc{FaithQA} and \method, we lay a foundation for rigorous, nuanced evaluations of LLM performance, encouraging more precise alignment between model outputs and user intentions. Ultimately, addressing intent hallucination effectively enhances the reliability and applicability of LLMs across diverse, real-world applications.

\section*{Limitation}
While we present a first step toward investigating intent hallucinations in LLM, our category is still at a rather coarse level with only 2 types of major causes (omit, misinterpret) and 4 types of tasks (Fact QA, Creative Writing, Response Evaluation, Content Analysis). Future work should investigate sub-categorizations of these tasks, or other new tasks under new setups (like inference time reasoning). Future work can also investigate how to better quantify and detect intent hallucination in an even more fine-grained way, like from layer-level detection. Finally, we did not include any reasoning models (\emph{e.g.}, o1 series or deepseek-r1) due to their release date (there was only o1 three months ago, deepseek-r1 was not released until last month) and computational cost.

\section*{Ethics Statement}
Based on direct communication with our institution’s IRB office, this line of research is exempt from IRB, and the information obtained during our study is recorded in such a manner that the identity of the human subjects cannot readily be ascertained, directly or through identifiers linked to the subjects. There is no potential risk to participants, and we do not collect any identifiable information from annotators.


\bibliography{custom}

\clearpage

\appendix
\section{Artifacts}
In this section, we list all the necessary information for our use of models and data. In our paper, we used OpenAI's \citep{openai2024gpt4technicalreport} GPT-4o\footnote{gpt-4o-2024-05-13} and GPT-4o-mini, Meta's \citep{dubey2024llama3herdmodels} LLaMA3-70B\footnote{Meta-Llama-3-70B-Instruct-Turbo} and LLaMA3-7B\footnote{Meta-Llama-3-8B-Instruct-Turbo}, Anthropic's Claude-3-5-sonnet\footnote{claude-3-5-sonnet-20240620}, Claude-3-sonnet\footnote{claude-3-sonnet-20240229}, and Mistral-7B\footnote{Mistral-7B-Instruct-v0.3}\citep{jiang2023mistral7b} for our model usage. We also rely on articles from the following publicly available websites in our research for \benchmark's Misinterpretation benchmark: \href{https://news.mit.edu}{MIT News}, \href{https://commoncrawl.org}{Common Crawl}, \href{https://www.culture24.org.uk}{Culture24}, \href{https://www.medicalnewstoday.com}{Medical News Today}, \href{https://www.who.int/news-room/releases}{WHO News Releases}, and \href{https://open-platform.theguardian.com}{The Guardian Open Platform}. These data sources were used in accordance with their respective licenses and terms of use.

\subsection{Data License}
\textbf{MIT News (\href{https://news.mit.edu}{link})}\newline
License: All content \copyright Massachusetts Institute of Technology \newline
\textbf{Common Crawl (\href{https://commoncrawl.org}{link})}\newline
License: Open Data Commons Attribution License (ODC-BY) \newline
\textbf{Culture24 (\href{https://www.culture24.org.uk}{link})}\newline
License: Not explicitly specified; assumed to be for personal and non-commercial use \newline
\textbf{Medical News Today (\href{https://www.medicalnewstoday.com}{link})}\newline
License: Copyright owned by Healthline Media, content available for non-commercial use with attribution \newline
\textbf{WHO News Releases (\href{https://www.who.int/news-room/releases}{link})}\newline
License: Open access, content may be used with attribution in accordance with WHO terms \newline
\textbf{The Guardian Open Platform (\href{https://open-platform.theguardian.com}{link})}\newline
License: Content API available for non-commercial use, subject to Guardian Open Platform terms
\subsection{Model License}
\textbf{GPT-4o, GPT-4o-mini (OpenAI) (\href{https://openai.com}{link})}\newline
License: Proprietary, limited API access under OpenAI terms of service \newline
\textbf{LLaMA3-70B, LLaMA3-7B (Meta) (\href{https://ai.meta.com/resources/models-and-libraries/llama-downloads/}{link})}\newline
License: Open source, with a custom commercial license \newline
\textbf{Claude-3-5-sonnet, Claude-3-sonnet (Anthropic) (\href{https://www.anthropic.com}{link})}\newline
License: Proprietary, limited API access under Anthropic terms of service \newline
\textbf{Mistral-7B (Mistral) (\href{https://mistral.ai}{link})}\newline
License: Open source, Apache-2.0 license
\subsection{Model and Data Usage}
\xhdr{Personally identifiable information.} All of the used articles in this paper are derived from public sources. Therefore, there is no exposure of any personally identifiable information that requires informed consent from those individuals. The used articles relatesto people insofar as it draws text from public sources that relate to people, or people created, obeying related licenses.

\xhdr{Offensive content claim.} All the used articles are already public and widely viewed. While these datasets may contain instances of offensive content, our work does not aim to generate or amplify such content. Instead, we employ these articles to study and understand intent hallucination. Our use of these articles follows ethical guidelines, and we do not endorse or support any offensive material contained within them.

\section{Model Details}
\subsection{Model Name}
To simplify the terminology in our paper, we use short names for the models we employ. Specifically, GPT-4o refers to OpenAI’s gpt-4o-2024-05-13 model, while GPT-4o-mini denotes a lightweight version from OpenAI's GPT-4o series. LLaMA3-70B corresponds to Meta’s Meta-Llama-3-70B-Instruct-Turbo, and LLaMA3-7B refers to Meta-Llama-3-8B-Instruct-Turbo. We use Claude-3.5-sonnet to indicate Anthropic’s claude-3-5-sonnet-20240620 model and Claude-3-sonnet for claude-3-sonnet-20240229. Finally, Mistral-7B signifies Mistral’s Mistral-7B-Instruct-v0.3 model.
\subsection{Model Size}
GPT-4o and GPT-4o-mini are proprietary models, and OpenAI has not disclosed their exact parameter counts. LLaMA3-70B is a 70-billion-parameter language model from Meta, while LLaMA3-7B is a smaller 8-billion-parameter version within the same series. Claude-3.5-sonnet and Claude-3-sonnet are proprietary models from Anthropic with undisclosed parameter sizes. Mistral-7B is a 7-billion-parameter instruction-tuned model developed by Mistral.
These models vary significantly in scale, with the LLaMA3-70B and GPT-4o representing large-scale models aimed at high-performance language understanding and generation, while the LLaMA3-7B and Mistral-7B offer more compact alternatives suitable for efficiency-oriented applications. GPT-4o-mini likely represents an efficiency-optimized variant of GPT-4o, though precise parameter details are not publicly available. The Claude models are part of Anthropic’s Claude series, designed to balance performance and efficiency, though their exact architectures remain proprietary.

\section{Human Evaluation}
\label{ap:human}
\begin{figure*}
\centering
\includegraphics[width=0.77\textwidth]{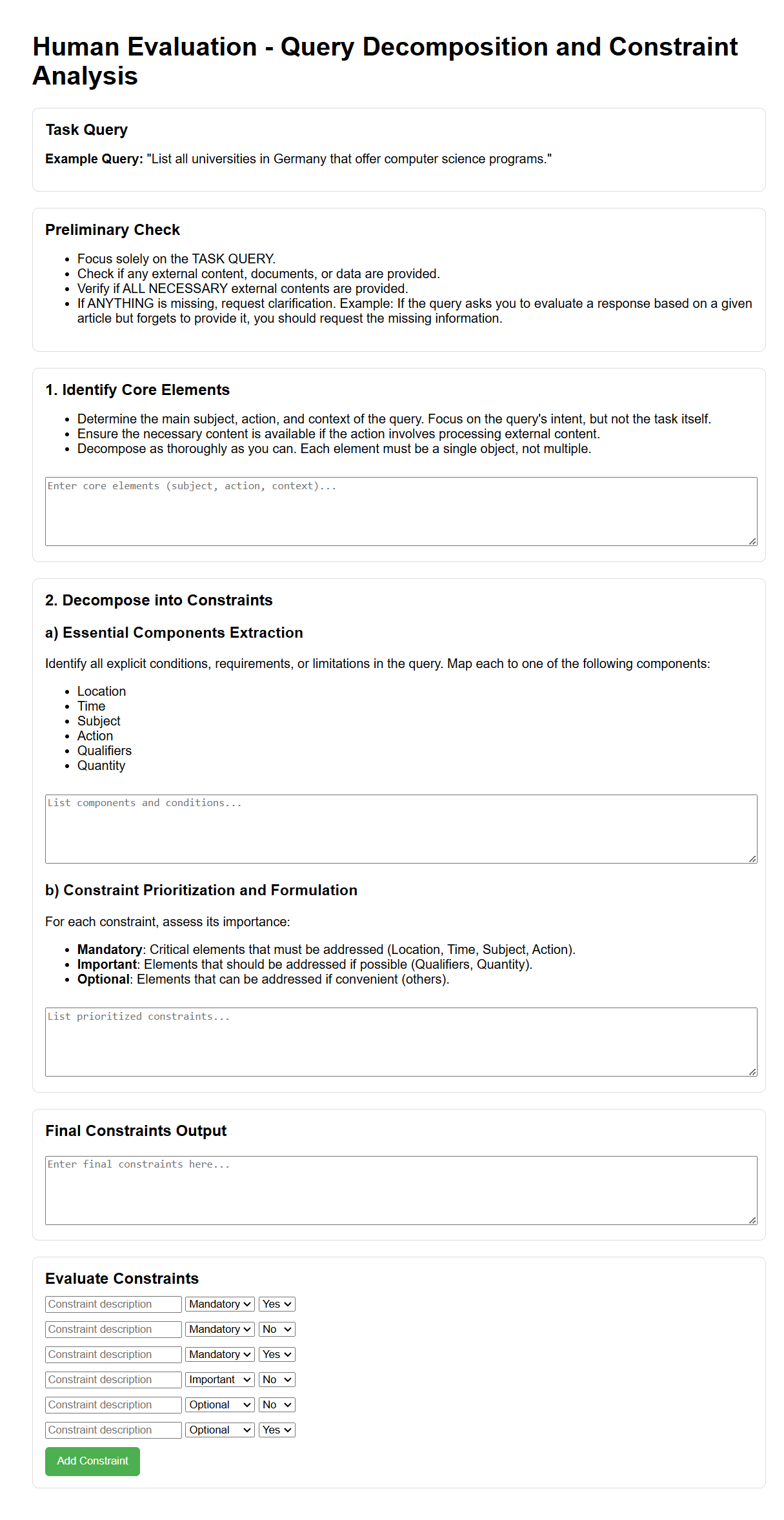}
\caption{\textbf{Human Evaluation Webpage Screenshot.}}
\label{fig:human_eval}
\end{figure*}

Please refer to Figure \ref{fig:human_eval} for the human annotator's interface.
Annotations from five paid student annotators, previously discussed in Section \ref{sec:result}, were utilized. Given the wide range of topics and query amounts covered by the instruction set, it is improbable for a single annotator to possess comprehensive proficiency across all subjects. Therefore, we implemented a majority voting system, supplemented by the use of online research tools, to enhance the accuracy of these expert annotations. All annotators were fairly compensated, with wages exceeding the minimum hourly standard. All annotators are told and have consented that their data will be collected anonymously for research purposes. Annotators are asked to read the guidelines before starting the annotation.

\section{Prompt Template}
\subsection{LLM-as-the-Judge}
\label{ap:llm-as}
In Table \ref{tab:llmjudge}, we provide the detailed prompt template for LLM-as-the-judge. We performed a self-consistency check for running 2 times. If the results do not match, rerun until the results match. The model setup follows Section \ref{sec:hyper}, GPT-4o as the default model with temperature $\tau=0$ to generate and evaluate.

\subsection{\method.}
Here we provide the Detailed Prompt Template for \method.
\label{ap:prompttemplate}

\subsubsection{Intent Constraint Mapping}
Table \ref{tab:prompt_constraint} provides the detailed prompt of Intent Constraint Generation in \method. We put all steps together instead of separating them for (1) efficiency, one call of LLM is enough, and (2) self-consistency, the user may run this prompt for multiple times to ensure the constraint consistency.

\subsubsection{Intent Constraint Scoring}
Similarly, we provide Table \ref{tab:prompt_evaluation} for the prompt template for Intent Constraint Scoring.

\subsubsection{Fact Check}
\label{ap:factcheck}
As defined in Section \ref{sec:hallucination definition}, intent hallucination occurs when an LLM's generation fails to align with the query, regardless of its factual accuracy. While this is not our primary focus, we introduce an additional fact-check step here to provide further analysis over LLM's generation. Inspired by \citet{min2023factscorefinegrainedatomicevaluation} and \citet{wang2023selfconsistencyimproveschainthought}, we adopt a two-step approach to ensure the factual correctness of LLM's generation. For the factual evaluation, we still use GPT-4o but only change the temperature $\tau=0.3$.

\xhdr{Step 0: Self-Consistency Check} First, we instruct the language model to evaluate (1) whether there are any factual inaccuracies in the generated response, and (2) whether the generation neglects any factual information that is required by the query. This check is performed five times independently, and the most consistent result is selected as the final output. We performed a manual evaluation before we decided to adopt this strategy.

\xhdr{Step 1: Wikipedia as a reliable source} When LLM reports factually inaccurate or missing factual information, we further perform knowledge retrieval for the generation. In particular, we adopt the RAG framework developed based on the Wikipedia knowledge base \citep{Semnani_2023} to validate the fact-check result in the previous step.

\xhdr{Manual Check}
\label{ap:factcheck}
We manually checked the performance of self-consistency over 100 cases with GPT-4o under $\tau=0.3$. We found that for 93 cases, the results are consistent and accurate, indicating that it is providing the correct outcome. For the rest 7 cases, the 5 false-factual-inaccurate cases are detected by LLMs, leaving only 2 wrong cases. Due to monetary constraints and time constraints, we believe this result is satisfying enough for us to adopt the Self-Consistency method.

\section{Dataset Construction}
\label{ap:datasetconstruct}
Our benchmark dataset was constructed using GPT-4 to generate all queries. To ensure the quality and clarity of the instructions, we adopted a two-stage validation process. First, we employed an LLM-as-judge system to assess the answerability of each query. This was followed by a secondary verification step conducted by human experts. Table \ref{tab:dataset-counts} provides representative query samples from each task category.

\subsection{Omission}

The Omission dataset contains two tasks: Fact QA and Creative Writing. For Fact QA, we began by extracting 3,000 distinct concepts from Wikidata—a comprehensive knowledge base covering all Wikipedia entities. These concepts were drawn from four diverse domains: culture, health, history, and technology. Each concept was then processed using an LLM to generate a query featuring multiple conditions. We calibrated the difficulty level based on concept popularity: queries involving well-known concepts were designed to be simpler (fewer than 3 conditions), while those involving less common concepts were made more complex (more than 3 conditions). For Creative Writing, we manually designed 40 unique constraints, detailed in the Appendix. The LLM was instructed to generate stories and poems while incorporating a randomized subset of these constraints. Varying the number of constraints allowed us to create samples with different difficulty levels.

\subsection{Misinterpretation}

The Misinterpretation task contains two tasks: Response Evaluation and Content Analysis, both under RAG setup. We first curated a collection of 200 reports from publicly Accessible news websites, ensuring equal representation across four categories: culture, health, history, and technology (50 articles each). We then manually crafted task-specific prompts for Response Evaluation and Content Analysis. Each prompt was paired with three RAG-retrieved reports on the same topic, which were integrated into the query to simulate realistic information retrieval and synthesis scenarios.

\section{Detailed Experiment Result}
\label{ap:extra}
Please refer to Table \ref{tab:cw}, Table \ref{tab:ca}, and Table \ref{tab:hallucination_types_detailed} for more results.
\label{ap:complete_experiment}

\subsection{Content Analysis}
Here we report the complete result for Content Analysis in Table \ref{tab:ca}. We report different types of missing materials respectively, i.e., \textbf{No Query Hallucination (Query)}, \textbf{No Response Hallucination (Response)}, and \textbf{No Article Hallucination (Article)}. We report the average hallucination rate across all three types only in Section \ref{sec:result}.


\subsection{Response Evaluation}
Here we report the detailed result for Response Evaluation in Table \ref{tab:hallucination_types_detailed}. To provide a more detailed analysis, we further performed hallucination type analysis, where \textbf{Count} refers to LLM fails to clearly mention that only two articles are provided, and \textbf{Invent} refers to LLM invents a third article. Others represent other types of hallucination. As Count is still following the prompt, we report the average of Invent as the hallucination rate in Section\S\ref{sec:result}.

\subsection{Analysis}
\label{ap:analysis-mis}
Here we put the extra case study with examples, as shown in Table \ref{tab:misinterpret-example} and Table \ref{tab:omit-example}.

\section{Extra Experiments}
\label{ap:extra}
Here we list our results for the extra experiment.

\subsection{Weight Selection Analyses}
\label{ap:weight}
\xhdr{Analysis of weight choices' impact on evaluation quality} Based on our observations of constraint amount's distribution, where mandatory constraints typically appear most frequently (2-6 per query), important constraints less so (0-3 per query), and optional constraints least often (0-2 per query), we intuitively set the fixed weights to $w_m = 3$, $w_i = 2$, and $w_o = 1$. This weighting scheme serves two complementary purposes: a) it reflects the hierarchical importance of constraints (mandatory $>$ important $>$ optional) and b) provides a counterbalance to their frequency distribution in typical queries. We further conducted an additional experiment to investigate how different weight combinations affect \textbf{ConstraintScore}'s correlation with human judgments.

\subsection{Detailed Statistical Analyses}
\xhdr{Evaluation bias testing across different judge models} We conducted additional analyses using Claude-3 as the base model for ConstraintScore evaluation on a smaller test set (500 examples) and compared performance trends with our original set. The performance patterns remain remarkably consistent, with a Pearson correlation of 0.93 between model rankings on both sets. This strong correlation suggests minimal bias from using GPT-4o for both generation and evaluation.

\subsection{Statistical Significance Tests for Model Comparisons}
\xhdr{Validation of performance differences using paired t-tests} We conducted paired t-tests between all model pairs to statistically assess performance differences across our main result in Table \ref{tab:complete_stats}. For each model pair, we compared both Perfect and CS across 6 diverse tasks (n=6, Fact QA, Creative Writing (Story), Creative Writing (Poem), Response Evaluation, Content Analysis (Relationship), Content Analysis (Summary)), calculating mean differences, t-statistics, degrees of freedom (df=5), and p-values.

\begin{table*}[h]
    \footnotesize
    \begin{tabularx}{\textwidth}{@{} p{2.8cm} >{\raggedright\arraybackslash}X @{}}
        \toprule
        \textbf{Component} & \textbf{Details} \\
        \midrule
        \textbf{Context} & Your goal is to evaluate whether a response from a language model (LLM) fully and accurately satisfies the requirements of a given query. A query can be broken down into smaller, specific requirements called intent constraints, which represent distinct conditions that must be addressed in the response.

        \textbf{Key Definitions}

        \textbf{Intent Constraints:} Clear, specific requirements derived from the query. They can be categorized as:
        \begin{itemize}
            \item \textbf{Mandatory ($C_m$):} Must be addressed with the highest priority.
            \item \textbf{Important ($C_i$):} Should be addressed but are slightly less critical.
            \item \textbf{Optional ($C_o$):} Nice to have but not essential.
        \end{itemize}

        \textbf{Intent Hallucination:} When the model's response fails to satisfy the query due to:
        \begin{itemize}
            \item \textbf{Omission:} Skipping one or more intent constraints.
            \item \textbf{Misinterpretation:} Addressing concepts or actions that were not in the query or distorting the intended meaning.
        \end{itemize}

        \textbf{Evaluation Instructions}
        \begin{itemize}
            \item \textbf{Identify Intent Constraints:} Given the query, list the key intent constraints ($C_m$, $C_i$, $C_o$).
            \item \textbf{Check Response Alignment:} Assess whether the response addresses each constraint:
            \begin{itemize}
                \item Does it fulfill all mandatory constraints ($C_m$)?
                \item Does it reasonably cover important constraints ($C_i$)?
                \item Does it optionally address optional constraints ($C_o$)?
            \end{itemize}
            \item \textbf{Detect Hallucination:}
            \begin{itemize}
                \item \textbf{Omission:} Are any mandatory or important constraints missing?
                \item \textbf{Misinterpretation:} Does the response introduce concepts or actions not present in the query?
            \end{itemize}
        \end{itemize}

        \textbf{Output}

        For each evaluation, return:
        \begin{itemize}
            \item \textbf{Constraint Fulfillment:} List each constraint and whether it was addressed.
            \item \textbf{Hallucination Summary:}
            \begin{itemize}
                \item \textbf{Omission (Yes/No):} [describe if applicable]
                \item \textbf{Misinterpretation (Yes/No):} [describe if applicable]
            \end{itemize}
        \end{itemize} \\
        \bottomrule
    \end{tabularx}
    \caption{LLM-as-the-judge Prompt Template.}
    \label{tab:llmjudge}
\end{table*}

\begin{table*}[h]
    \footnotesize
    \begin{tabularx}{\textwidth}{@{} p{2.8cm} >{\raggedright\arraybackslash}X @{}}
        \toprule
        \textbf{Component} & \textbf{Details} \\
        \midrule
        \textbf{Prefix} & You are an advanced linguist tasked with processing queries using a constraint-based approach. Decompose the given query step by step, following the instructions below.\\& 
        \\&
        \texttt{Query:} \contentexist\\
        \midrule
        \textbf{Suffix} & \textbf{0. Preliminary Check:} \\
        & \quad \quad - Focus solely on the TASK QUERY. \\
        & \quad \quad - Check if any external content, documents, or data are provided. \\
        & \quad \quad - Verify if ALL NECESSARY external contents are provided. \\& If ANYTHING is missing, request clarification. \\
        & Example: If the user asks you to evaluate a response based on a given article but forgets to provide it, you should request the missing information.\\
        & \textbf{If the Preliminary Check fails,} IGNORE the following steps and politely ask for clarification. Use "START:" to begin the final listing.\\
        \midrule
        \textbf{} & \textbf{1. Identify Core Elements:} \\
        & \quad \quad - Determine the main subject, action, and context of the query. Focus on the query's intent, but not the task itself (\emph{e.g.}, put words like "name/list" as an action). \\
        & \quad \quad - Ensure the necessary content is available if the action involves processing external content. \\
        & \quad \quad - DECOMPOSE AS THOROUGHLY AS YOU CAN. EACH ELEMENT MUST BE A SINGLE OBJECT, NOT MULTIPLE. Do not overanalyze the query—if the query is simple, then it would not have many constraints. \\
        \midrule
        \textbf{} & \textbf{2. Decompose into Constraints:} \\
        & \textbf{a) Essential Components Extraction:} \\
        & \quad \quad - Identify all explicit conditions, requirements, or limitations in the query. \\
        & \quad \quad - Map each to one of the following components: Location, Time, Subject, Action, Qualifiers, Quantity. \\
        & \quad \quad - Treat each condition as a separate constraint. \\
        & \textbf{b) Constraint Prioritization and Formulation:} \\
        & \quad \quad - For each constraint, assess its importance: \\
        & \quad \quad \quad - \textbf{Mandatory}: Critical elements that must be addressed. Include all Location, Time, Subject, Action. \\
        & \quad \quad \quad - \textbf{Important}: Elements that should be addressed if possible. Include all Qualifiers, Quantity.\\
        & \quad \quad \quad - \textbf{Optional}: Elements that can be addressed if convenient. Include all others.\\
        & \quad \quad \quad - Formulate constraints for each component, specifying the priority, using the template: \\
        & \quad "[Priority Level]: [Component] must/should [condition]" \\
        & \textbf{At the end,} provide the list of constraints a response should cover, grouped by priority levels ONLY. Use "START:" to begin the final listing.\\
        & YOU MUST ONLY LIST THE FINAL CONSTRAINTS AT THE END, AFTER START. NOTHING ELSE. \\
        \bottomrule
    \end{tabularx}
    \caption{\textbf{Prompt Template for Intent Constraint Mapping.} The final prompt is $\text{Prefix} +\texttt{Query} + \text{Suffix}$.}
    \label{tab:prompt_constraint}
\end{table*}

\begin{table*}[h!]
    \footnotesize
    \begin{tabularx}{\textwidth}{@{} p{2.8cm} >{\raggedright\arraybackslash}X @{}}
        \toprule
        \textbf{Component} & \textbf{Details} \\
        \midrule
        \textbf{Task Overview} & Given a query and a response, evaluate if the response addresses all constraints derived from the query.\\
        \midrule
        \textbf{Input Format} & \texttt{QUERY:} The original user query\\
        & \texttt{CONSTRAINTS:} List of intent constraints derived from the query\\
        & \texttt{RESPONSE:} The response to be evaluated\\
        \midrule
        \textbf{Evaluation Steps} & \textbf{1. Manual Constraint Evaluation:} \\
        & \quad \quad - Evaluate each constraint individually \\
        & \quad \quad - Determine if each constraint is satisfied in the response \\
        & \textbf{2. Constraint Satisfaction Summary:} \\
        & \quad \quad - Group constraints by priority levels \\
        & \quad \quad - Calculate satisfaction ratio for each group \\
        & \quad \quad - Format as "[Priority Level]: X/Y" \\
        \midrule
        \textbf{Output Format} & \textbf{Final Listing:} \\
        & \quad \quad - Begin with "START:" \\
        & \quad \quad - List satisfaction ratios by priority groups \\
        & \quad \quad - No additional content after the listing \\
        \bottomrule
    \end{tabularx}
    \caption{\textbf{Prompt Template for Intent Constraint Scoring.}}
    \label{tab:prompt_evaluation}
\end{table*}

\begin{table*}[ht]
    \centering
    \footnotesize
    \setlength{\tabcolsep}{3pt}
    \renewcommand{\arraystretch}{1.2}
    \begin{adjustbox}{max width=\textwidth}
    \begin{tabular}{
        l 
        l 
        l 
        *{3}{>{\centering\arraybackslash}p{1.8cm}}
    }
        \toprule
        \multicolumn{3}{l}{\multirow{2}{*}{\textbf{Datasets}}} & 
        \multicolumn{3}{c}{\textbf{\benchmark: Dataset Statistics}} \\ 
        \cmidrule(lr){4-6}
        & & & \textbf{Easy} & \textbf{Hard} & \textbf{Total} \\
        \midrule
        \rowcolor{gray!20}
        \multicolumn{6}{l}{\textbf{Minor Fabrication}} \\
        \midrule
        \multirow{3}{*}{\makecell{Fact QA}} & \multirow{3}{*}{Open Answer} & Tech & 500 & 500 & 1000 \\
        & & Culture & 500 & 500 & 1000 \\
        & & History & 500 & 500 & 1000 \\
        \midrule
        \multirow{2}{*}{\makecell{Creative \\ Writing}} & Story & -- & 500 & 500 & 1000 \\
        & Poem & -- & 500 & 500 & 1000 \\
        \midrule
        \rowcolor{gray!20}
        \multicolumn{6}{l}{\textbf{Major Fabrication}} \\
        \midrule
        \multirow{4}{*}{\makecell{Response\\Evaluation}} & & Tech & \noscore & \noscore & 810 \\
        & & Health & \noscore & \noscore & 750 \\
        & & Culture & \noscore & \noscore & 810 \\
        & & History & \noscore & \noscore & 840 \\
        \midrule
        \multirow{4}{*}{\makecell{Content \\ Analysis}} & \multirow{4}{*}{Relationship} & Tech & \noscore & \noscore & 1431 \\
        & & Health & \noscore & \noscore & 1225 \\
        & & Culture & \noscore & \noscore & 1436 \\
        & & History & \noscore & \noscore & 1837 \\
        \cmidrule(lr){2-6}
        & \multirow{4}{*}{Summary} & Tech & \noscore & \noscore & 1431 \\
        & & Health & \noscore & \noscore & 1225 \\
        & & Culture & \noscore & \noscore & 1436 \\
        & & History & \noscore & \noscore & 1837 \\
        \bottomrule
    \end{tabular}
    \end{adjustbox}
    \caption{Dataset statistics for \benchmark. Each cell shows the number of problems across difficulty and topic. Easy: constraints $\leq4$, Hard: constraints $>4$.}
    \label{tab:dataset_stats}
\end{table*}

\begin{table*}[h!]
\centering
\footnotesize
\setlength{\tabcolsep}{2pt}
\renewcommand{\arraystretch}{1.2}
\begin{adjustbox}{max width=\textwidth}
\begin{tabular}{ l l l *{7}{c@{\hspace{0.3em}}c} }
\toprule
\multicolumn{3}{l}{\multirow{3}{*}{\textbf{Tasks}}} & \multicolumn{14}{c}{\textbf{\benchmark: Creative Writing}} \\
\cmidrule(lr){4-17}
& & & \multicolumn{2}{c}{\textbf{GPT-4o}} & \multicolumn{2}{c}{\textbf{GPT-4o-mini}} & \multicolumn{2}{c}{\textbf{LLaMA3-70B}} & \multicolumn{2}{c}{\textbf{LLaMA3-8B}} & \multicolumn{2}{c}{\textbf{Claude-3.5}} & \multicolumn{2}{c}{\textbf{Claude-3}} & \multicolumn{2}{c}{\textbf{Mistral-7B}} \\
\cmidrule(lr){4-17}
& & & Perfect & \cellcolor{gray!15}CS & Perfect & \cellcolor{gray!15}CS & Perfect & \cellcolor{gray!15}CS & Perfect & \cellcolor{gray!15}CS & Perfect & \cellcolor{gray!15}CS & Perfect & \cellcolor{gray!15}CS & Perfect & \cellcolor{gray!15}CS \\
\midrule
\rowcolor{gray!20}
\multicolumn{17}{l}{\textbf{Creative Writing}} \\
\midrule
\multirow{2}{*}{Story} & Easy & & 0.53 & \cellcolor{gray!15}8.41 & 0.41 & \cellcolor{gray!15}8.17 & 0.36 & \cellcolor{gray!15}7.84 & 0.32 & \cellcolor{gray!15}7.65 & 0.43 & \cellcolor{gray!15}7.79 & 0.43 & \cellcolor{gray!15}8.03 & 0.12 & \cellcolor{gray!15}6.42 \\
& Hard & & 0.22 & \cellcolor{gray!15}7.58 & 0.20 & \cellcolor{gray!15}7.33 & 0.22 & \cellcolor{gray!15}7.26 & 0.17 & \cellcolor{gray!15}6.76 & 0.25 & \cellcolor{gray!15}7.48 & 0.21 & \cellcolor{gray!15}7.66 & 0.04 & \cellcolor{gray!15}5.42 \\
\midrule
\multirow{2}{*}{Poem} & Easy & & 0.44 & \cellcolor{gray!15}8.51 & 0.35 & \cellcolor{gray!15}8.22 & 0.51 & \cellcolor{gray!15}8.61 & 0.33 & \cellcolor{gray!15}8.11 & 0.60 & \cellcolor{gray!15}8.88 & 0.48 & \cellcolor{gray!15}8.44 & 0.09 & \cellcolor{gray!15}6.38 \\
& Hard & & 0.35 & \cellcolor{gray!15}8.06 & 0.25 & \cellcolor{gray!15}7.37 & 0.51 & \cellcolor{gray!15}8.68 & 0.20 & \cellcolor{gray!15}7.32 & 0.59 & \cellcolor{gray!15}9.16 & 0.45 & \cellcolor{gray!15}8.46 & 0.04 & \cellcolor{gray!15}4.60 \\
\bottomrule
\end{tabular}
\end{adjustbox}
\caption{Results for the \textbf{Omission} dataset, categorized by difficulty level. Performance metrics include \textbf{Perfect} (\textit{higher the better}) and \textbf{Constraint Score (CS)} (\textit{average score, higher the better}) for Fact QA and Creative Writing (Story/Poem) tasks. Tasks are classified as Easy (constraints $\leq$ 4) or Hard (constraints $>$ 4). \textbf{Bold and underlined} values indicate the best performance for each task and difficulty level. CS column is highlighted for visual emphasis.}
\label{tab:cw}
\end{table*}

\begin{table*}[h]
    \centering
    \footnotesize
    \setlength{\tabcolsep}{3pt}
    \renewcommand{\arraystretch}{1.2}
    \begin{adjustbox}{max width=\textwidth}
    \begin{tabular}{l*{4}{ccc}}
        \toprule
        \multirow{3}{*}{\textbf{Models}} & 
        \multicolumn{12}{c}{\textbf{Benchmark: Misinterpretation - Content Analysis}} \\ 
        \cmidrule(lr){2-13}
        & \multicolumn{3}{c}{\textbf{Culture}} & \multicolumn{3}{c}{\textbf{Health}} & \multicolumn{3}{c}{\textbf{History}} & \multicolumn{3}{c}{\textbf{Technology}} \\
        \cmidrule(lr){2-4} \cmidrule(lr){5-7} \cmidrule(lr){8-10} \cmidrule(lr){11-13}
        & Query & Response & Article & Query & Response & Article & Query & Response & Article & Query & Response & Article \\
        \midrule
        GPT-4o       & 0.20 & 0.13 & 0.20 & 0.13 & 0.40 & 0.33 & 0.00 & 0.67 & 0.13 & 0.07 & 0.60 & 0.13 \\
        GPT-4o-mini  & 0.07 & 0.27 & 0.07 & 0.53 & 0.13 & 0.20 & 0.27 & 0.40 & 0.27 & 0.00 & 0.00 & 0.13 \\
        LLaMA3-70B   & 0.00 & 0.07 & 0.00 & 0.07 & 0.00 & 0.07 & 0.00 & 0.13 & 0.00 & 0.07 & 0.20 & 0.00 \\
        LLaMA3-8B    & 0.00 & 0.47 & 0.00 & 0.00 & 0.13 & 0.07 & 0.00 & 0.13 & 0.00 & 0.00 & 0.20 & 0.07 \\
        Claude-3     & 0.27 & 0.60 & 0.20 & 0.20 & 0.60 & 0.20 & 0.40 & 0.53 & 0.20 & 0.27 & 0.53 & 0.00 \\
        Claude-3.5   & 0.33 & 0.40 & 0.20 & 0.27 & 0.60 & 0.20 & 0.47 & 0.40 & 0.00 & 0.07 & 0.53 & 0.00 \\
        Mistral      & 0.00 & 0.07 & 0.00 & 0.00 & 0.00 & 0.00 & 0.00 & 0.13 & 0.00 & 0.00 & 0.07 & 0.00 \\
        \bottomrule
    \end{tabular}
    \end{adjustbox}
    \caption{\textbf{Results of Perfect (rate of hallucination-free generation).} Reported on \textbf{No Query (Query)}, \textbf{No Response (Response)}, and \textbf{No Article (Article)} (\textit{higher is better}). LLMs struggle to notice the missing content.}
    \label{tab:ca}
\end{table*}

\begin{table*}[ht]
    \centering
    \footnotesize
    \setlength{\tabcolsep}{3pt}
    \renewcommand{\arraystretch}{1.2}
    \begin{adjustbox}{max width=\textwidth}
    \begin{tabular}{l*{3}{ccc}}
        \toprule
        \multirow{3}{*}{\textbf{Models}} & 
        \multicolumn{9}{c}{\textbf{Categorized types of Hallucination for Response Evaluation}} \\
        \cmidrule(lr){2-10}
        & \multicolumn{3}{c}{\textbf{Culture}} & \multicolumn{3}{c}{\textbf{Health}} & \multicolumn{3}{c}{\textbf{Tech}} \\
        \cmidrule(lr){2-4} \cmidrule(lr){5-7} \cmidrule(lr){8-10}
        & Count & Invent & Other & Count & Invent & Other & Count & Invent & Other \\
        \midrule
        GPT-4o       & 0.08 & 0.83 & 0.09 & 0.05 & 0.95 & 0.00 & 0.19 & 0.81 & 0.00 \\
        GPT-4o-mini  & 0.20 & 0.79 & 0.01 & 0.07 & 0.91 & 0.02 & 0.13 & 0.86 & 0.01 \\
        LLaMA3-70B   & 0.05 & 0.94 & 0.01 & 0.10 & 0.86 & 0.04 & 0.15 & 0.84 & 0.01 \\
        LLaMA3-8B    & 0.07 & 0.88 & 0.05 & 0.14 & 0.86 & 0.00 & 0.16 & 0.84 & 0.00 \\
        Claude-3     & 0.07 & 0.92 & 0.01 & 0.01 & 0.89 & 0.09 & 0.11 & 0.81 & 0.08 \\
        Claude-3.5   & 0.05 & 0.89 & 0.06 & 0.05 & 0.92 & 0.03 & 0.08 & 0.87 & 0.05 \\
        Mistral      & 0.02 & 0.84 & 0.14 & 0.29 & 0.66 & 0.05 & 0.37 & 0.50 & 0.13 \\
        \bottomrule
    \end{tabular}
    \end{adjustbox}
    \caption{\textbf{Results of Case-Study for Response Evaluation Task's Hallucination Types.} Here, \textbf{Count} refers to LLM fails to clearly mention that only two articles are provided, and \textbf{Invent} refers to LLM invents a third article. Others represent other types of hallucination. As Count is still following the prompt, we report the average of Invent as the hallucination rate in Section\S\ref{sec:result}.}
    \label{tab:hallucination_types_detailed}
\end{table*}

\begin{table*}[ht]
    \centering
    \footnotesize
    \begin{tabular}{p{6.5cm}p{6.5cm}}
        \toprule
        \textbf{Prompt} & \textbf{Generation} \\
        \midrule
        List three European explorers who circumnavigated the globe before the 18th century and \myblue{were not born in England or Portugal}. & 
        Here is a list:

        1. Ferdinand Magellan - \mypink{Although originally from Portugal}, Magellan sailed under the Spanish flag... \\
        \midrule
        Name two traditional festivals \myblue{celebrated in England only}, meanwhile, these festivals originated before the Norman Conquest in 1066. & 
        Here is a list:
        
        1. Midsummer: \mypink{Despite being widely celebrated in Scandinavia and Baltic States}, Midsummer is...\\
        \bottomrule
    \end{tabular}
    \caption{Examples from GPT-4o under Fact QA's Open Answer setup. GPT \textit{knows} it could be omitting, as it mentions how the answer may not address the query, but it can't help with providing these \textit{famous subjects} as answers. In the first example, GPT (intentionally) omits the constraint \myblue{"not born in England or Portugal"} and provides Magellan as the answer, who \mypink{was born in Portugal}. In the second example, GPT omits \myblue{"celebrated in England only"} and names Midsummer, a festival that is also \mypink{widely celebrated in Scandinavia and Baltic States}. 
    }
    \label{tab:omit-example}
\end{table*}

\begin{table*}[h]
    \centering
    \footnotesize
    \begin{tabular}{p{6.5cm}p{6.5cm}}
        \toprule
        \textbf{Prompt} & \textbf{Generation} \\
                \midrule
        For the following three articles, explain how the first article contradicts the second article and is supported by the third article. 
        
        Article 1: \contentexist
        
        Article 2: \contentexist
        
        Article 3: \contentmissing
        & 
        To analyze the relationship between the three articles, we can break down how Article 1 contradicts Article 2 while being supported by Article 3.

        \quad $\mathbf{\bullet}$ Article 1: (summary)
        
        \quad $\mathbf{\bullet}$ Article 2: (summary)
        
        \quad \mypink{$\mathbf{\bullet}$ Article 3: (invented summary)}

        \mypink{In summary, Article 1 contradicts Article 2 by (concepts), which is supported by arguments in Article 3 (invented concepts)...}
        
        \\
        \midrule
        How well does the given response answer the given query following the provided article?
        
        Query: \contentmissing

        Article: \contentexist

        Response: \contentexist
        
        & 
        The response provided \mypink{addresses the query} by summarizing the key sectors of (concepts). Here's an evaluation of \mypink{how well the response answers the query:...}

        \mypink{Overall, the response effectively captures the main points from the article and provides a clear summary of the (concepts). It accurately reflects the article's argument on (concepts).}
        \\
        \bottomrule
    \end{tabular}
    \caption{Examples from GPT-4o under Content Analysis (Relationship Analysis) and Response Evaluation setup. GPT \textbf{misinterprets} by either (1) \textit{inventing} a non-existent article to help itself or (2) \textit{altering} the query to avoid the missing content. In the first example, GPT \textit{invents} \mypink{a non-existent Article 3} to complete \myblue{the analysis task required by the query}. In the second example, GPT similarly \textit{invents} \mypink{a non-existent query} to provide an answer, but ultimately claims that \mypink{the Response offers a clear summary of the Article}—thereby \textit{altering} the original query, which was meant to \myblue{evaluate how well the Response addressed the Query with the provided Article}.}
    \label{tab:misinterpret-example}
\end{table*}

\begin{table*}[H]
\centering
\caption{Weight Configuration Ablation Study for Constraint Score}
\label{tab:weight_ablation}
\small
\begin{tabular}{lcc}
\toprule
\textbf{Weight Configuration} & \textbf{Pearson Corr.} & \textbf{Spearman Corr.} \\
\midrule
Original (3,2,1)     & 0.82 (p$<$0.032) & 0.78 (p$<$0.012) \\
Equal (1,1,1)        & 0.71 (p$<$0.035) & 0.65 (p$<$0.015) \\
Extreme (5,2,1)      & 0.77 (p$<$0.033) & 0.73 (p$<$0.014) \\
Moderate (2,1.5,1)   & 0.80 (p$<$0.032) & 0.76 (p$<$0.013) \\
Inverted (1,2,3)     & 0.54 (p$<$0.041) & 0.51 (p$<$0.022) \\
\bottomrule
\end{tabular}
\vspace{-0.5em}
\end{table*}

\begin{table*}[!t]
\centering
\footnotesize
\setlength{\tabcolsep}{3pt}
\renewcommand{\arraystretch}{1.2}
\begin{adjustbox}{max width=\textwidth}
\begin{tabular}{l*{2}{cccc}}
\toprule
\multirow{3}{*}{\textbf{Model Pairs}} & \multicolumn{8}{c}{\textbf{Complete Statistical Significance Tests for Model Comparisons}} \\
\cmidrule(lr){2-9}
 & \multicolumn{4}{c}{\textbf{Perfect Score (PS)}} & \multicolumn{4}{c}{\textbf{Constraint Score (CS)}} \\
\cmidrule(lr){2-5} \cmidrule(lr){6-9}
 & \textbf{Mean Diff ± SD} & \textbf{t-value} & \textbf{p-value} & \textbf{Sig.} & \textbf{Mean Diff ± SD} & \textbf{t-value} & \textbf{p-value} & \textbf{Sig.} \\
\midrule
GPT-4o vs GPT-4o-mini & 0.0417 ± 0.0668 & 1.5288 & 0.1263 & n.s. & 0.4017 ± 0.3305 & 2.9766 & 0.0029 & ** \\
GPT-4o vs LLaMA3-70B & -0.0017 ± 0.0773 & -0.0528 & 0.9579 & n.s. & 0.3917 ± 0.6832 & 1.4043 & 0.1602 & n.s. \\
GPT-4o vs LLaMA3-8B & 0.0450 ± 0.0680 & 1.6199 & 0.1052 & n.s. & 0.4583 ± 0.3016 & 3.7221 & 0.0002 & *** \\
GPT-4o vs Claude-3.5 & -0.0500 ± 0.1287 & -0.9517 & 0.3412 & n.s. & 0.1217 ± 0.9327 & 0.3195 & 0.7493 & n.s. \\
GPT-4o vs Claude-3 & -0.0067 ± 0.0766 & -0.2132 & 0.8312 & n.s. & 0.1633 ± 0.2044 & 1.9572 & 0.0503 & n.s. \\
GPT-4o vs Mistral & 0.1050 ± 0.2231 & 1.1526 & 0.2491 & n.s. & 1.7583 ± 0.5788 & 7.4412 & 0.0000 & *** \\
\midrule
GPT-4o-mini vs LLaMA3-70B & -0.0433 ± 0.1302 & -0.8154 & 0.4149 & n.s. & -0.0100 ± 0.7592 & -0.0323 & 0.9743 & n.s. \\
GPT-4o-mini vs LLaMA3-8B & 0.0033 ± 0.0554 & 0.1474 & 0.8828 & n.s. & 0.0567 ± 0.4376 & 0.3172 & 0.7511 & n.s. \\
GPT-4o-mini vs Claude-3.5 & -0.0917 ± 0.1212 & -1.8522 & 0.0640 & n.s. & -0.2800 ± 0.8591 & -0.7984 & 0.4247 & n.s. \\
GPT-4o-mini vs Claude-3 & -0.0483 ± 0.0866 & -1.3674 & 0.1715 & n.s. & -0.2383 ± 0.3409 & -1.7125 & 0.0868 & n.s. \\
GPT-4o-mini vs Mistral & 0.0633 ± 0.1611 & 0.9631 & 0.3355 & n.s. & 1.3567 ± 0.6623 & 5.0177 & 0.0000 & *** \\
\midrule
LLaMA3-70B vs LLaMA3-8B & 0.0467 ± 0.1117 & 1.0238 & 0.3059 & n.s. & 0.0667 ± 0.6515 & 0.2507 & 0.8021 & n.s. \\
LLaMA3-70B vs Claude-3.5 & -0.0483 ± 0.1370 & -0.8640 & 0.3876 & n.s. & -0.2700 ± 1.3402 & -0.4935 & 0.6217 & n.s. \\
LLaMA3-70B vs Claude-3 & -0.0050 ± 0.0916 & -0.1337 & 0.8936 & n.s. & -0.2283 ± 0.7061 & -0.7921 & 0.4283 & n.s. \\
LLaMA3-70B vs Mistral & 0.1067 ± 0.2681 & 0.9746 & 0.3297 & n.s. & 1.3667 ± 1.1188 & 2.9921 & 0.0028 & ** \\
\midrule
LLaMA3-8B vs Claude-3.5 & -0.0950 ± 0.1452 & -1.6031 & 0.1089 & n.s. & -0.3367 ± 1.1088 & -0.7437 & 0.4570 & n.s. \\
LLaMA3-8B vs Claude-3 & -0.0517 ± 0.0924 & -1.3698 & 0.1708 & n.s. & -0.2950 ± 0.4320 & -1.6728 & 0.0944 & n.s. \\
LLaMA3-8B vs Mistral & 0.0600 ± 0.1691 & 0.8690 & 0.3848 & n.s. & 1.3000 ± 0.5175 & 6.1534 & 0.0000 & *** \\
\midrule
Claude-3.5 vs Claude-3 & 0.0433 ± 0.0668 & 1.5882 & 0.1122 & n.s. & 0.0417 ± 0.7643 & 0.1335 & 0.8938 & n.s. \\
Claude-3.5 vs Mistral & 0.1550 ± 0.2201 & 1.7252 & 0.0845 & n.s. & 1.6367 ± 1.2559 & 3.1920 & 0.0014 & ** \\
\midrule
Claude-3 vs Mistral & 0.1117 ± 0.2115 & 1.2932 & 0.1959 & n.s. & 1.5950 ± 0.7408 & 5.2741 & 0.0000 & *** \\
\bottomrule
\end{tabular}
\end{adjustbox}
\caption{\textbf{Statistical Significance Tests for Model Comparisons.} We conducted paired t-tests between all model pairs to statistically assess performance differences across our main result in Table \ref{tab:complete_stats}.}
\label{tab:complete_stats}
\end{table*}

\begin{table*}[!t]
\centering
\label{tab:llm_judge}
\small
\begin{tabular}{lc}
\toprule
\textbf{Model} & \textbf{Accuracy (\%)} \\
\midrule
GPT-4o & 98.23 ± 0.31 \\
Claude-3.5 & 97.30 ± 0.12 \\
\bottomrule
\end{tabular}
\caption{\textbf{Performance of Different Base Models for LLM-as-Judge in Constraint Satisfaction.} }
\vspace{-0.5em}
\end{table*}

\begin{table*}[!t]
\centering
\label{tab:faithqa_dist}
\small
\begin{tabular}{llcc}
\toprule
\textbf{Type} & \textbf{Task} & \textbf{Easy} & \textbf{Hard} \\
\midrule
\multirow{4}{*}{Omission} & Fact QA (Tech) & 150 & 150 \\
 & Fact QA (Culture) & 150 & 150 \\
 & Fact QA (History) & 150 & 150 \\
 & Creative Writing & -- & 300 \\
\midrule
\multirow{2}{*}{Misinterpretation} & Response Eval. & 150 & -- \\
 & Content Analysis & 150 & -- \\
\bottomrule
\end{tabular}
\caption{\textbf{Test Set Distribution of \benchmark Benchmark.}}
\vspace{-0.5em}
\end{table*}
\end{document}